%% file: Formatting-Instructions-LaTeX-2025.tex
\documentclass[letterpaper]{article} 
\usepackage{aaai25}  
\usepackage{times}  
\usepackage{helvet}  
\usepackage{courier}  
\usepackage[hyphens]{url}  
\usepackage{graphicx} 
\usepackage{natbib}  
\usepackage{caption} 
\usepackage{algorithm}
\usepackage{algorithmic}
\usepackage[table]{xcolor}
\usepackage{tabularx}  
\usepackage{amsmath} 
\usepackage{amsthm}  
\usepackage{graphicx}
\usepackage{booktabs} 
\usepackage{enumitem}
\usepackage{bm}
\newtheorem{theorem}{Theorem}  
  
\usepackage{amssymb}  
\usepackage{multirow}
\input{math_commands.tex}

\usepackage{newfloat}
\usepackage{listings}
\DeclareCaptionStyle{ruled}{labelfont=normalfont,labelsep=colon,strut=off} 
\floatstyle{ruled}
\newfloat{listing}{tb}{lst}{}
\floatname{listing}{Listing}
\title{Disentangling Long-Short Term State Under Unknown Interventions for Online Time Series Forecasting}
\author{
  Ruichu Cai\textsuperscript{\rm1, \rm2},
  Haiqin Huang\textsuperscript{\rm1},
  Zhifan
  Jiang\textsuperscript{\rm1},
  Zijian Li\textsuperscript{\rm3}\footnote{corresponding author: Zijian Li (leizigin@gmail.com)},
  Changze Zhou\textsuperscript{\rm1},\\
  Yuequn Liu\textsuperscript{\rm1},
  Yuming Liu\textsuperscript{\rm1},
  Zhifeng Hao\textsuperscript{\rm4}
}
\affiliations{
    \textsuperscript{\rm 1}School of Computer Science, Guangdong University of Technology, China\\
    \textsuperscript{\rm 2}Peng Cheng Laboratory, Shenzhen, China\\
    \textsuperscript{\rm 3}Machine Learning Department, Mohamed bin Zayed University of Artificial Intelligence, United Arab Emirates\\
     \textsuperscript{\rm 4}Shantou University\\

}

\usepackage{bibentry}

\begin{document}

\maketitle

\begin{abstract}
Current methods for time series forecasting struggle in the online scenario, since it is difficult to preserve long-term dependency while adapting short-term changes when data are arriving sequentially. Although some recent methods solve this problem by controlling the updates of latent states, they cannot disentangle the long/short-term states, leading to the inability to effectively adapt to nonstationary. To tackle this challenge, we propose a general framework to disentangle long/short-term states for online time series forecasting. Our idea is inspired by the observations where short-term changes can be led by unknown interventions like abrupt policies in the stock market. Based on this insight, we formalize a data generation process with unknown interventions on short-term states. Under mild assumptions, we further leverage the independence of short-term states led by unknown interventions to establish the identification theory to achieve the disentanglement of long/short-term states. Built on this theory, we develop a \textbf{L}ong \textbf{S}hort-\textbf{T}erm \textbf{D}isentanglement model (\textbf{LSTD}) to extract the long/short-term states with long/short term encoders, respectively. Furthermore, the \textbf{LSTD} model incorporates a smooth constraint to preserve the long-term dependencies and an interrupted dependency constraint to enforce the forgetting of short-term dependencies, together boosting the disentanglement of long/short-term states. Experimental results on several benchmark datasets show that our \textbf{LSTD} model outperforms existing methods for online time series forecasting, validating its efficacy in real-world applications.
\end{abstract}
\begin{links}
    \link{Code}{https://github.com/DMIRLAB-Group/LSTD}
\end{links}
\begin{figure*}[t]
\centering
\includegraphics[width=1.9\columnwidth]{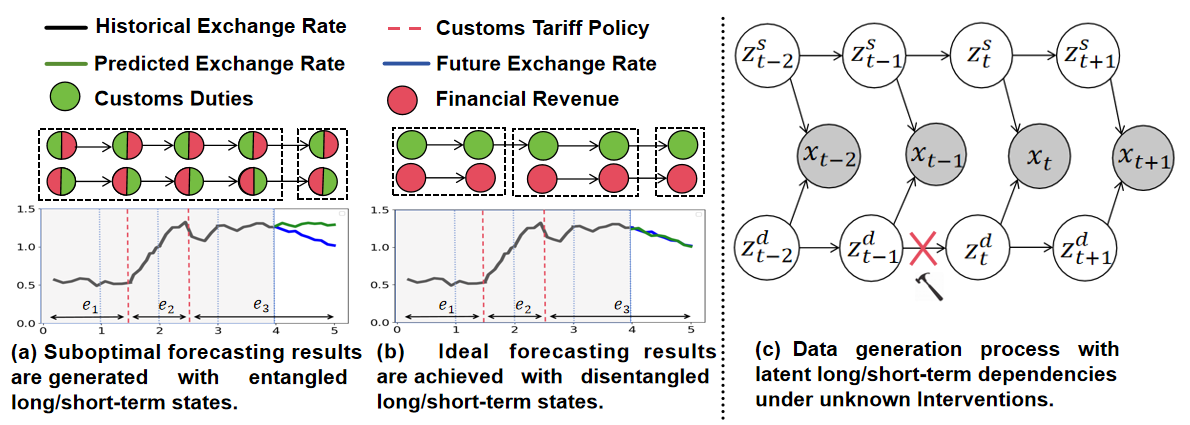}
    \caption{Illustration of sequentially arriving exchange rate data, which is influenced by short-term customs duties and long-term financial revenue. Moreover, the short-term customs duties are intervened by sudden customs tariff policies. (a) If the estimated short-term customs duties and long-term financial revenue are entangled, short-term influence from Environments (e.g., $e_1, e_2, e_3$) may affect the effectiveness of the models to adapt to the changing environments, leading to suboptimal forecasting performance. (b) When the long/short-term states are disentangled, the model can quickly adapt to environmental changes and hence achieve correct forecasting results. (c) Data generation process for time-series data. $\bm{z}_{t}^s$ and $\bm{z}_{t}^d$ denotes the long/short-term states. Note that the short-term states $\bm{z}_{t}^d$ are intervened randomly.}
    \label{fig:motivation}
\end{figure*}

\section{Introduction}


As one of the most fundamental tasks in time series analysis \cite{hamilton2020time,liu2023itransformer}, time series forecasting \cite{zhou2021informer,zeng2023transformers,kitaev2020reformer,liu2021pyraformer,wu2021autoformer,zhou2021informer} plays a critical role in various fields such as finance \cite{clements2004forecasting,cao2019financial}, and traffic \cite{lippi2013short}. However, in the industry, since time series data often arrives sequentially and is accompanied by temporal distribution shifts \cite{wang2022koopman,li2024and}, existing methods \cite{wu2021autoformer,nie2022time,lopez2017gradient} that heavily rely on the mini-batch training paradigm can hardly adapt to these changing distributions, leading to suboptimal prediction results in the online scenario.


To solve this problem, several recent methodologies \cite{cai2023memda,guo2024online,mejri2024novel,lin1992self} are proposed to adapt the short-term nonstationarity and long-term dependencies. FSNet \cite{pham2022learning} leverages the partial derivative to characterize the short-term information and an associative memory to preserve the long-term dependencies. To better combine long-term and short-term historical information, OneNet \cite{wen2024onenet} uses a reinforcement learning-based to dynamically adjust the combination of temporal correlation and cross-variable dependency models. Recently, Zhang et al. \cite{zhang2024addressing} propose the  Concept Drift Detection and Adaptation framework (D$^3$A), which first detects the temporal distribution shift and then employs an aggressive manner to update the model. In summary, these methods aim to address online time series forecasting via two steps: 1) disentangling long/short-term states; and 2) adapting short-term states and reusing long-term states for forecasting. Please refer to Appendix A for further discussion about online time series forecasting and causal representation learning.



Although current methods achieve non-trivial contributions on how to update short-term states or how to efficiently combine the long/short-term states, 
they implicitly assume that the long/short-term states have been well-disentangled from nonstationary time series data. However, this assumption is hard to meet, and without disentanglement of the long/short-term states, existing methods can hardly adapt to the nonstationary environments. 
Figure \ref{fig:motivation} provides a finance example, where the monetary exchange rate is influenced by the short-term variables (e.g. customs duties) and the long-term variables (e.g., financial revenue). Nonstationarity occurs due to unknown customs tariff policies. 
As shown in Figure \ref{fig:motivation} (a), when the long-term and short-term latent variables are not disentangled, the financial revenues are entangled with the customs duties. As a result, existing methods can be hard to effectively adapt to the changes in financial revenues and may obtain an inaccurate forecasting performance even if they use a masterly strategy to update the short-term states and preserve the long-term states. 



Based on the aforementioned example, we observe that nonstationarity is brought by the unknown interventions on short-term states. Moreover, to address the online forecasting task, it is intuitive to find that we should disentangle the long/short-term states from the time series with unknown interventions as shown in Figure \ref{fig:motivation} (b). Under this intuition, we first consider that the sequentially arriving data follow a data generation process in Figure \ref{fig:motivation} (c), where the latent short-term states are influenced by unknown interventions. Under mild assumptions, we establish disentanglement results on long/short-term states by leveraging the independence of intervened short-term states. To bridge the gap between theory and practice, we further develop a \textbf{L}ong \textbf{S}hort-\textbf{T}erm \textbf{D}isentanglement model (\textbf{LSTD}) to solve the online time series forecasting problem. Specifically, the proposed \textbf{LSTD} model includes a minimal update constraint to preserve the long-term dependencies and an interrupted dependency constraint to enforce the forgetting of short-term dependencies, which facilitates the disentanglement of long-term and short-term latent states.  Empirical results on several real-world benchmark datasets show that the proposed \textbf{LSTD} method outperforms existing state-of-the-art methods for online time series forecasting, highlighting its effectiveness in real-world applications.



\section{Data Generation Process for Time Series Data}

To show how to disentangle the long-term and short-term latent states in the online time series forecasting scenario, we first introduce the data generation process of time series data as shown in Figure \ref{fig:motivation} (c). Mathematically, we let $\rvx=\{\rvx_1,\rvx_2,\cdots,\rvx_t,\cdots\}$ be time series data with discrete time steps, in which each observation $\rvx_t$ is generated from latent variables $\rvz_t$ through an invertible and nonlinear mixing function $g$ as formalized in Equation (\ref{equ:mixing}).
\begin{equation}
\label{equ:mixing}
    \rvx_t=g(\rvz_t).
\end{equation}

At each time step $t$, $\rvz_t \in \mathbb{R}^{n}$ are divided into the long-term latent states $\rvz_t^s \in \mathbb{R}^{n_s}$ and short-term latent states $\rvz_t^d \in \mathbb{R}^{n_d}$, and $n=n_s+n_d$. Moreover, the $i$-th component of $\rvz_t^s$ is generated by some components of historical long-term latent states $\rvz_{t-\tau}^s$ with the time lag of $\tau$ via a nonparametric function as shown in Equation (\ref{equ:long_term_gen}).
\begin{equation}
\label{equ:long_term_gen}
\begin{split}
    z^s_{t,i}=f^s_i(\{z^s_{t-\tau,k}|& z^s_{t-\tau,k}\in \textbf{Pa}(z^s_{t,i})\}, \varepsilon^s_{t,i}), \\
    &\text{with}\quad \varepsilon^s_{t,i} \sim p_{\varepsilon^s_{t,i}},
\end{split}
\end{equation}
where $\textbf{Pa}(z^s_{t,i})$ denotes the set of latent variables that directly cause $z^s_{t,i}$ and $\varepsilon^s_{t,i}$ denotes the temporally
and spatially independent noise extracted from a distribution $p_{\varepsilon^s_{t,i}}$.

Moreover, according to the observation in the example in Figure \ref{fig:motivation}, we assume that the nonstationarity in time series data is led by the interventions on the short-term latent variables (e.g., the truncation between $\rvz_{t-1}^d$ and $\rvz_t^d$ in Figure \ref{fig:motivation} (c)). It is noted that when the interventions occur is unknown. To illustrate the randomness of the interventions, we let $I$ be an indicator to decide if an intervention occurs and $I$ comes from a Bernoulli distribution $\mathbf{B}(I, \theta )$ with the probability of $\theta $. When $I=0$, it indicates no intervention, whereas when $I=1$, it signifies intervention. When intervention occurs, the data is generated solely by noise. Formally, the generation process of the short-term latent variables is shown as follows:
\begin{equation}
\label{equ:short_term_gen}
\begin{split}
    z_{t,j}^d&=\left\{ 
    \begin{aligned}
         f^d_j(\{z^d_{t-\tau,k}|& z^d_{t-\tau,k}\in \textbf{Pa}(z^d_{t,j})\}, \varepsilon^d_{t,j}) \;\;\text{, if I = 0}   \\
        &f^d_j(\varepsilon^d_{t,j}) \quad\quad\quad\quad\quad\qquad,\text{if I = 1}
    \end{aligned}
    \right.\\
    &\quad\quad\text{where}\quad \varepsilon^d_{t,j} \sim p_{\varepsilon^d_{t,j}} \quad \text{and} \quad I\sim \mathbf{B}(I, \theta),
\end{split}
\end{equation}
where $\textbf{Pa}(z^d_{t,j})$ denotes the set of latent variables that directly cause $z^d_{t,j}$ and $\varepsilon^d_{t,j}$ denotes the temporally and spatially independent noise extracted from a distribution $p_{\varepsilon^d_{t,j}}$. 

The data generation process as shown in Equation (\ref{equ:mixing})-(\ref{equ:short_term_gen}) can be well interpreted by the aforementioned financial example. First, the exchange rate can be considered as the observation time series data. Sequentially, the financial revenue and the customs duties denote the long-term and short-term latent variables, respectively. Finally, $I=1$ denotes that the customs tariff policy intervenes with customs duties and leads to temporal distribution changes. In the context of online time series forecasting, where the time series data arrive sequentially, we first predict the value of $\rvx_{L+1:H}$ given $\rvx_{1:L}$ at $t$-th time step. Then at $t+1$-th time step, we have access to the true value of $\rvx_{L+1:H}$ to update the model and then use $\rvx_{2:L+1}$ to predict the value of $\rvx_{L+2:H+1}$.

\section{Disentanglement of Long-Term and Short-Term States}

To disentangle the long-term latent variables $\textbf{z}^s_t$ and the short-term latent variables $\textbf{z}^d_t$, we propose the block-wise identification theory in Theory \ref{theorem:1}. Mathematically, the block-wise identification means that for the ground-truth $\rvz_t^*$, there exists $\hat{\rvz}_t^*$ and an invertible function $h^*_z:\mathbb{R}^{n^*} \rightarrow \mathbb{R}^{n^*}$, such that $\hat{\rvz}_t^*=h^*_z(\rvz_t^*)$. And $*$ can be $d$ or $s$.
\begin{theorem}
\label{theorem:1}
(\textbf{Subspace Identification of the long-term and short-term Latent Variables}) Suppose that the observed data from long/short-term is generated following the data generation process in Figure\ref{fig:motivation} (c), and we further make the following assumptions:
\begin{itemize}[leftmargin=*,  itemsep=5pt]  
    \item A1 \underline{(\textbf{Smooth, Positive and Conditional independent Den-}} \\
    \underline{\textbf{sity:})}
    \cite{yao2022temporally,yao2021learning} The probability density function of latent variables is smooth and positive, i.e., $p(\rvz_{t-\tau+1:t}|\rvz_{t-\tau})>0$ over $\textbf{Z}_{t-\tau}$ and $\textbf{Z}_{t-\tau+1:t}$.
    Conditioned on $\rvz_{t-\tau}$ each $z_i$ is independent of any other $z_j$ for $i,j\in {1,...,n}, i\neq j,\ i.e$, $\log{p(\rvz_{t-\tau+1:t}|\rvz_{t-\tau})} = \sum_{k=1}^{n_s} \log {p(z_{t-\tau+1:t,k}|\rvz_{t-\tau})}$
    \item A2 \underline{(\textbf{non-singular Jacobian}):} \cite{kong2023understanding} Each generating function $g$ has non-singular Jacobian matrices almost anywhere and $g$ is invertible. 
    \item A3\underline{ (\textbf{Linear Independence}:) }\cite{yao2022temporally} For any $\rvz^d\in \textbf{Z}^d_{t-\tau+1:t}\subseteq {R}^{n_d},\bar{v}_{t-\tau,1},...,\bar{v}_{t-\tau,n_d}$ as $n_d$ vector functions in $z^d_{t-\tau,1},...,z^d_{t-\tau.l},...,z^d_{t-\tau,n_d}$ are linear independent, where $\bar{v}_{t-\tau,l}$ are formalized as follows:
     \begin{align}
        \bar{\textbf{v}}_{t-\tau,l}=\frac{\partial^2 \log{p(\rvz^d_{t-\tau+1:t}|\rvz^d_{t-\tau})}  }{\partial z^d_{t-\tau+1:t ,k} \partial z^d_{t-\tau,l} } 
    \end{align}
\end{itemize}
Suppose that we learn $(\hat{g}, {\hat{f}}_i^s, {\hat{f}}_i^d)$ to achieve Equation (\ref{equ:mixing})-(\ref{equ:short_term_gen}) with the minimal number of transition edge among short term latent variables $\rvz^d_1, \cdots, \rvz^d_t, \cdots$, then the long-term and short-term latent variables are block-wise identifiable.

\end{theorem}

\paragraph{Proof Sketch:} The proof can be found in Appendix \textbf{B}. First, we construct an invertible transformation $h_z$ between the ground-truth latent variables and estimated ones. Sequentially, we prove that the ground truth of long-term latent variables is not the function of short-term latent variables by leveraging the pairing time series from different influences. Sequentially, we leverage sufficient variability of historical information to show that the short-term latent variables are not the function of the estimated long-term latent variables. Moreover, by leveraging the invertibility of transformation $h_z$, we can obtain the Jacobian of $h_z$ as shown in Equation (\ref{equ:Jh1}), where $B = 0$ and $C = 0$, since the ground truth long-term latent variables are not the functions of short-term latent variables and the short-term latent variables are not the function of the estimated long-term latent variables.
     
\begin{equation}  
    \begin{gathered}  
    \label{equ:Jh1}  
        \textbf{J}_{h_z}=\begin{bmatrix}  
        \begin{array}{c|c}  
            \textbf{A}:=\frac{\partial \rvz_t^{s}}{\partial \hat{\rvz}_t^{s}} & \textbf{B}:=\frac{\partial \rvz_t^s}{\partial \hat{\rvz}_t^{d}}=0 \\ \midrule  
            \textbf{C}:=\frac{\partial \rvz_t^{d}}{\partial \hat{\rvz}_t^s}=0 & \textbf{D}:=\frac{\partial \rvz_t^{d}}{\partial \hat{\rvz}_t^{d}}
        \end{array}  
        \end{bmatrix}  
    \end{gathered}  
    \end{equation} 

\paragraph{Discussion of the Identification Results:}
We would like to highlight that the theoretical results provide sufficient conditions for the identification of our model. That implies: 1) our model can be correctly identified when all the assumptions hold. 2) at the same time, even if some of the above assumptions do not hold, our method may still learn the correct model. From an application perspective, these assumptions rigorously defined a subset of applicable scenarios of our model. Thus, we provide detailed explanations of the assumptions, how they relate to real-world scenarios, and in which scenarios they are satisfied. 

\paragraph{Smooth, Positive and Conditional independent Density.} 
This assumption is common in the existing identification results \cite{yao2022temporally,yao2021learning,yao2022temporally,yao2021learning}. In real-world scenarios, smooth and positive density implies continuous changes in historical information, such as temperature variations in weather data. To achieve this, we should sample as much data as possible to learn the transition probabilities more accurately. Moreover, The conditional independent assumption is also common in identifying temporal latent processes  \cite{li2024subspace}. Intuitively, it means there are no immediate relations among latent variables. To satisfy this assumption, we can sample data at high frequency to avoid instantaneous dependencies caused by subsampling. 

\paragraph{Non-singular Jacobian of $g$.} This assumption is also common in \cite{kong2023understanding, li2024and, li2024identification,xie2023multi,kong2023identification}. Mathematically, it denotes that the Jacobian from the latent variables to the observed variables is full rank. 
In real-world scenarios, it means that there is at least one observation for each latent variable. To meet this assumption, we can ignore such independent latent variables since they have no influence on the observations. 


\paragraph{Linear Independence.}

\paragraph{Data generation process with unknown interventions.} In real-world time series data, there are many unknown interventions that lead to nonstationarity like the financial example in Figure 1. Therefore, this assumption is reasonable. Besides, we need to impose discontinuities in the short-term components to break the symmetry between the long and short terms in the causal graph. This ensures that the long and short terms are identifiable. Through the identifiable theory, we can explain whether the module learns long-term or short-term components, thereby theoretically guaranteeing the disentanglement of long and short terms. In practice, we may investigate the nonstationarity of the data to test whether this assumption is valid. 
\section{Long Short-Term Disentanglement Model}
\begin{figure*}[t]
    \centering
\includegraphics[width=1.8\columnwidth]{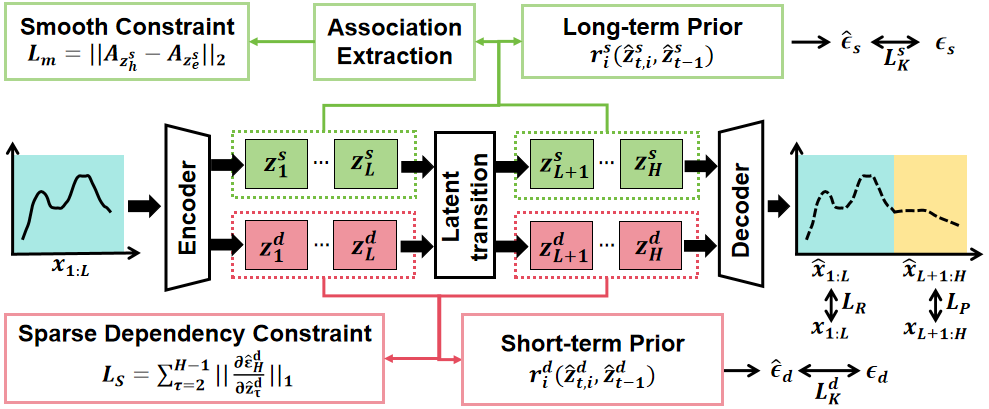}
    \caption{The framework of the proposed \textbf{LSTD} model. The long/short-term latent variables $\rvz_{1:L}^d$ and $\rvz_{1:L}^s$ are extracted from the encoder. And the latent transition module is used to estimated the $\rvz_{L+1:H}^d$ and the $\rvz_{L+1:H}^s$ from $\rvz_{1:L}^d$ and $\rvz_{1:L}^s$, respectively. The long-term and short-term prior networks are used to estimate the prior distributions.} 
    \label{fig:model}
\end{figure*}
\subsection{Model Overview}
In this section, we introduce the implementation of the long/short-term disentanglement model as shown in Figure \ref{fig:model}. Specifically, it uses a variational sequential autoencoder as a backbone architecture and further employs long-term and short-term prior architectures with smooth constraint and sparse dependency constraint for long-term and short-term latent variable disentanglement. 


\subsection{Variational Sequential Autoencoder}
To model the time series data, we follow the data generation process in Figure \ref{fig:motivation} (c) and derive the evidence lower bound (ELBO) as shown in Equation (\ref{equ:elbo}).
\begin{equation}
\small
\label{equ:elbo}
\begin{split}
ELBO=&\underbrace{\mathbb{E}_{q(\rvz^s_{1:H}|\rvx_{1:H})} \mathbb{E}_{q(\rvz^d_{1:H}|\rvx_{1:H})}\ln p(\rvx_{1:H}|\rvz^s_{1:H},\rvz^d_{1:H})}_{L_{R}+L_{P}}\\&- \underbrace{D_{KL}(q(\rvz^s_{1:H}|\rvx_{1:H})||p(\rvz^s_{1:H}))}_{L^s_{K}}\\&
\underbrace{-D_{KL}(q(\rvz^d_{1:H}|\rvx_{1:H})||p(\rvz^d_{1:H}))}_{L^d_{K}}
\end{split}
\end{equation}
where $L_R$ and $L_P$ denote the reconstructed and prediction loss, respectively:
\begin{equation}
\small
    \begin{split}
    L_R&=\frac{1}{L} \sum^L_{i=1} (\hat{\rvz}_i-\rvz_i)^2
    \\L_P&=\frac{1}{H-L} \sum^H_{i=L+1} (\hat{\rvz}_i-\rvz_i)^2
    \end{split}
\end{equation}
$D_{KL}$ denotes the KL divergence. Specifically, $q(\rvz^s_{1:H}|\rvx_{1:H}),\ q(\rvz^d_{1:H}|\rvx_{1:H})$, which includes the encoder and the latent transition module in Figure \ref{fig:model}, is used to approximate the prior distribution. $p(\rvx_{1:H}|\rvz^s_{1:H},\rvz^d_{1:H})$ is used to reconstruct the historical observations and forecast the future values. The aforementioned two distributions can be formalized as follows:
\begin{equation}
\small
    \hat{\rvz}^s_{1:H},\hat{\rvz}^d_{1:H}=\phi(\rvx_{1:H}), \quad \hat{\rvx}_{1:H}=\psi(\rvz_{1:H}),
\end{equation}
Where $\rvz_{1:H}$ denotes the combination of $\hat{\rvz}^s_{1:H}$ and $\hat{\rvz}^d_{1:H}$. For the implementation of $\phi$, we follow the backbone of FSNet \cite{pham2022learning}. For the implementation of $\psi$, we employ an MLP (Multilayer Perceptron). Please refer to Appendix C for more implementation details of the \textbf{LSTD} model.

\subsection{Long-Term and Short-Term Prior Networks 
}
To model the prior distribution of the long-term latent variables, we propose the long-term prior networks. Similar to the existing methods for causal representation learning \cite{yao2021learning,yao2022temporally}, we let $\{r_i^s\}$ be a set of learned inverse transition functions that take the estimated long-term latent variables and output the noise term, i.e.,
$\hat{\epsilon}_{t,i}^s=r_i^s(\hat{z}_{t,i}^s, \hat{\rvz}^s_{t-1})$ \footnote{We use the superscript symbol to denote estimated variables.} and each $r_i^s$ is modeled with MLPs. Then we devise a transformation $\kappa^s:=\{\hat{\rvz}^s_{t-1},\hat{\rvz}^s_t\}\rightarrow\{\hat{\rvz}^s_{t-1}, \hat{\epsilon}^s_t\}$, and its Jacobian is $\small{\mathbf{J}_{\kappa^s}=
    \begin{pmatrix}
        \mathbb{I}&0\\
        M & \text{diag}\left(\frac{\partial r^s_i}{\partial \hat{z}_{t,i}^s}\right)
    \end{pmatrix}}$,
where $M$ denotes a matrix. By applying the change of variables formula, we have the following equation:
\begin{equation}
\label{equ:short_temp_transition}
    \log p(\hat{\rvz}^s_{t-1},\hat{\rvz}^s_t)=\log p(\hat{\rvz}^s_{t-1},\hat{\epsilon}_t^s) + \log|\text{det}(\mathbf{J}_{\kappa^s})|.
\end{equation}
Since we assume that the noise term in Equation (\ref{equ:short_temp_transition}) is independent with $\rvz_{t-1}^s$, we can enforce the independence of the estimated noise $\hat{\epsilon}_t^s$ and further have:
\begin{equation}
\log p(\hat{\rvz}^s_{t}|\hat{\rvz}^s_{t-1})=\log p(\hat{\epsilon}^s_{t}) + \sum_{i=1}^{n_s}\log|\frac{\partial r_i^s}{\partial \hat{z}^s_{t,i}}|.
\end{equation}
Therefore, the long-term prior can be estimated as follows:
\begin{equation}
\small
\log p(\hat{\rvz}_{1:t}^s)=\log p(\hat{\rvz}^s_1)+\sum_{\tau=2}^t \left( \sum_{i=1}^{n_s}\log p(\hat{\epsilon}^s_{\tau,i}) +\sum_{i=1}^{n_s} \log|\frac{\partial r_i^s}{\partial \hat{z}^s_{\tau,i}}| \right),
\end{equation}
where $p(\hat{\epsilon}^s_i)$ follow Gaussian distributions. Similarly, we can further estimate the short-term prior as follows:
\begin{equation}
\small
\log p(\hat{\rvz}_{1:t}^d)=\log p(\hat{\rvz}^d_1)+\sum_{\tau=2}^t \left( \sum_{i=1}^{n_d}\log p(\hat{\epsilon}^d_{\tau,i}) +\sum_{i=1}^{n_d} \log|\frac{\partial r_i^d}{\partial \hat{z}^d_{\tau,i}}| \right),
\end{equation}
\subsection{Smooth Constraint for Long-Term Disentanglement}
To preserve the long-term dependencies in the long-term latent variables, we propose the smooth constraint. Since the causal relationships of the long-term dependencies are stable, the association of the long-term dependencies is also stable. Based on this insight, we consider the attention weights as associations and extract the association with the help of the self-attention mechanism. Specifically, we first split the $\rvz_{1:H}^s$ into two equal-size segmentation $\rvz_{1:H/2}^s$ and $\rvz_{H/2:H}^s$. And then the association of $\rvz_{1:H/2}^s$ and $\rvz_{H/2:H}^s$ can be formalized as follows:
\begin{equation}
\begin{split}
    A_{\rvz_h^s}&=\text{Softmax}(\frac{\rvz_{1:H/2}^s { \rvz_{1:H/2}^s}^{\mathsf{T}}}{\sqrt{n_s}}),\\
    A_{\rvz_e^s}&=\text{Softmax}(\frac{\rvz_{H/2:H}^s { \rvz_{H/2:H}^s}^{\mathsf{T}}}{\sqrt{n_s}}),
\end{split}
\end{equation}
in which $A_{\rvz_h^s}$ and $A_{\rvz_e^s}$ denote the association matrices of the start half and the end half segments. Hence, we can restrict the long-term dependencies by restricting the similarity of these two matrices as shown in Equation (\ref{equ:smooth_con})
\begin{equation}
\label{equ:smooth_con}
    \mathcal{L}_m = ||A_{\rvz_h^s} - A_{\rvz_e^s}||_2,
\end{equation}
where $||\cdot||_2$ denotes the L2 norm of matrices.

\subsection{Interrupted Dependency Constraint for Short-Term Disentanglement}
Since the nonstationarity is assumed to be led by the interventions to the short-term latent variables, given $\rvz_{1:H}^d$, if intervention occurs at $\tau$-th time step, and $2<\tau<H-1$, then $\frac{\partial \varepsilon_{H, i}^d}{\partial z_{\tau-1,j}^d}=0$, where $i,j \in \{1,\cdots,n_d\}$. Based on this intuition, we aim to enforce the interruption of the estimated short-term dependencies to meet the unknown interventions. To achieve this, we propose the interrupted dependency constraint for the short-term variables. Specifically, given the estimated short-term variables $\rvz_{1:H}^d$, we have:
\begin{equation}
    \mathcal{L}_s = \sum_{(i,j)\in \{1,\cdots,n_d\}}\sum_{\tau\in\{2,\cdots,H-1\}}||\frac{\partial \hat{\varepsilon}_{H, i}^d}{\partial \hat{z}_{\tau-1,j}^d}||_1,
\end{equation}
where $||\cdot||_1$ denote the L1 norm.

By using the aforementioned interrupted dependency constraint, the intervention on the short-term latent variables can be automatically detected, which finally enforces the disentanglement of the short-term latent variables.

\subsection{Model Summary}
By combining the aforementioned variational sequentially autoencoder with the restriction of smooth constraint and interrupted dependency constraint, we can finally formalize the total loss of the proposed \textbf{LSTD} model as follows:
\begin{equation}
    \mathcal{L} =L_{R} + L_{P} + \beta L_K + \alpha L_m + \gamma L_s,
\end{equation}
where $L_{K}=L^s_{K}+L^d_{K}$ . And
$\alpha,\beta,\gamma$ are hyper-parameters.

\section{Experiment}

\subsection{Datasets}
To evaluate the performance of our method, we consider the following datasets. \textbf{ETT}  is an electricity transformer temperature dataset collected from two separate counties in China, which contains two separate datasets $\{\text{ETTh2, ETTm1}\}$ for one hour level and minutes level, respectively. \textbf{Exchange} is the daily exchange rate dataset from eight foreign countries including Australia, British, Canada, Switzerland, China, Japan, New Zealand, and Singapore ranging from 1990 to 2016. \textbf{Weather }\footnote{\url{https://www.bgc-jena.mpg.de/wetter/}} is recorded at the Weather Station at the Max Planck Institute
for Biogeochemistry in Jena, Germany. \textbf{ECL} \footnote{\url{https://archive.ics.uci.edu/dataset/321/electricityloaddiagrams20112014}} is an electricity-consuming load dataset with the electricity consumption (kWh) collected from 321 clients. \textbf{Traffic} \footnote{\url{https://pems.dot.ca.gov/}} is a dataset of traffic speeds collected from the California Transportation Agencies (CalTrans) Performance Measurement System (PeMS). For each dataset, we follow the standard preprocessing and setting in OneNet \cite{wen2024onenet}.

\subsection{Baselines}



\begin{table*}[]

\centering

\renewcommand{\arraystretch}{0.65}
\setlength{\tabcolsep}{4pt}

\begin{tabular}{c|c|cccccccccc}
\toprule
Models                     & Len & LSTD                                  & OneNet         & FSNet & OneNet-T       & DER++ & ER    & MIR   & TFCL   & Online-T & Informer \\ \midrule
                           & 1   & \textbf{0.377}                        & 0.380          & 0.466 & 0.411          & 0.508 & 0.508 & 0.486 & 0.557  & 0.502    & 7.571    \\
                           & 24  & 0.543                        & \textbf{0.532} & 0.687 & 0.772          & 0.828 & 0.808 & 0.812 & 0.846  & 0.830    & 4.629    \\
\multirow{-3}{*}{ETTh2}    & 48  & 0.616                        & \textbf{0.609} & 0.846 & 0.806          & 1.157 & 1.136 & 1.103 & 1.208  & 1.183    & 5.692    \\ \midrule
                           & 1   & \textbf{0.081}                        & 0.082 & 0.085 & 
                           0.082 & 0.083 & 0.086 & 0.085 & 0.087  & 0.214    & 0.456    \\
                           & 24  & 0.102                       & \textbf{0.098} & 0.115 & 0.212          & 0.196 & 0.202 & 0.192 & 0.211  & 0.258    & 0.478    \\
\multirow{-3}{*}{ETTm1}    & 48  & 0.115                                 & \textbf{0.108} & 0.127 & 0.223          & 0.208 & 0.220 & 0.210 & 0.236  & 0.283    & 0.388    \\ \midrule
                           & 1   & \textbf{0.153}                        & 0.156          & 0.162 & 0.171          & 0.174 & 0.180 & 0.179 & 0.177  & 0.206    & 0.426    \\
                           & 24  & \textbf{0.136}                        & 0.175          & 0.188 & 0.293          & 0.287 & 0.293 & 0.291 & 0.301  & 0.308    & 0.380    \\
\multirow{-3}{*}{WTH}      & 48  & \textbf{0.157}                        & 0.200          & 0.223 & 0.310          & 0.294 & 0.297 & 0.297 & 0.323  & 0.302    & 0.367    \\ \midrule
                           & 1   & \textbf{2.112}                        & 2.351          & 3.143 & 2.470          & 2.657 & 2.579 & 2.575 & 2.732  & 3.309    & 3.813    \\
                           & 24  & \textbf{1.422}                        & 2.074          & 6.051 & 4.713          & 8.996 & 9.327 & 9.265 & 12.094 & 11.339   & 9.185    \\
\multirow{-3}{*}{ECL}      & 48  & \textbf{1.411}                        & 2.201          & 7.034 & 4.567          & 9.009 & 9.685 & 9.411 & 12.110 & 11.534   & 11.183   \\ \midrule
                           & 1   & \textbf{0.231}                        & 0.241          & 0.312 & 0.236          & 0.271 & 0.284 & 0.298 & 0.306  & 0.334    & 0.234    \\
                           & 24  & \textbf{0.398}                        & 0.438          & 0.426 & 0.425          & 0.476 & 0.461 & 0.451 & 0.441  & 0.481    & 0.451    \\
\multirow{-3}{*}{Traffic}  & 48  & \textbf{0.426}                        & 0.473          & 0.445 & 0.451          & 0.486 & 0.510 & 0.502 & 0.438  & 0.503    & 0.496    \\ \midrule
                           & 1   & {\color[HTML]{212121} \textbf{0.013}} & 0.017          & 0.094 & 0.031          & 0.106 & 0.097 & 0.095 & 0.106  & 0.113    & 0.102    \\
                           & 24  & \textbf{0.039}                        & 0.047          & 0.113 & 0.060          & 0.111 & 0.162 & 0.104 & 0.098  & 0.116    & 0.107    \\
\multirow{-3}{*}{Exchange} & 48  & \textbf{0.043}                        & 0.062          & 0.156 & 0.065          & 0.183 & 0.181 & 0.101 & 0.101  & 0.168    & 0.116    \\ \bottomrule
\end{tabular}%
\caption{Mean Square Error (MSE) results on the different datasets. TCN is abbreviated as T}
\label{MSE}
\end{table*}

We consider nine state-of-the-art as follows: OneNet \cite{wen2024onenet} which considered the temporal and feature relationships and used reinforcement learning to update their relationships in real-time. At the same time, we compared with a very excellent backbone model FSNet \cite{pham2022learning} which considered gradient updates to optimize fast new
as well as retained information and be used in OneNet. Besides, we also compared the OneNet model with TCN as its backbone named OnetNet-TCN, and the regular usage of TCN named Online-TCN \cite{zinkevich2003online} for online learning. The Experience Replay (ER) \cite{chaudhry2019tiny} stored the previous data in a buffer and interleaved with newer samples during learning. Meanwhile, ER has many advanced variants: TFCL \cite{aljundi2019task} used a task-boundary detection
mechanism and a knowledge consolidation strategy; MIR \cite{aljundi2019online} selected
samples that cause the most forgetting; and DER++ \cite{buzzega2020dark} incorporated a
knowledge distillation strategy. Additionally, we have incorporated a long-term time-series forecasting model, Informer \cite{zhou2021informer}, to investigate the performance of conventional forecasting models in online time-series forecasting problems.
\subsection{Quantitative Results and Discussion}
 Experiment results on each dataset are shown in Table \ref{MSE} and Table \ref{MAE}. Since some methods report the best results on the original paper, we also show
the best results on the aforementioned tables.  Please refer to Appendix D for the experiment results with mean and variance over three random seeds. Our \textbf{LSTD} model significantly outperforms all other baselines on most online forecasting tasks. Specifically, our method outperforms the most
competitive baselines by a clear margin of $44\%$ on the Exchange, which verifies the example in the introduction. Moreover, our method also greatly reduced prediction errors in the WTH and ECL datasets. However,
our method achieves the second-best but still comparable
results in the ETT dataset, this might be because there are a few unknown interventions in the ETT datasets. How to address other types of nonstationarity will be an interesting future direction. In addition, we conduct performance analysis experiments and visualization in Appendix D. Compared with other models, we can find that the proposed \textbf{LSTD} has the best model performance and relatively good model efficiency.
\begin{figure}
    \centering
    \includegraphics[width=1\linewidth]{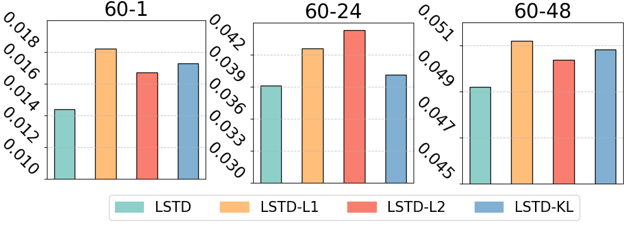}
    \caption{ Ablation study on the Exchange datasets. We explore the impact of different loss terms}
    \label{fig:ablation}
\end{figure}

\subsection{Qualitative Results and Discussion}
\begin{figure*}[t]
    \centering
    \includegraphics[width=0.9\linewidth]
    {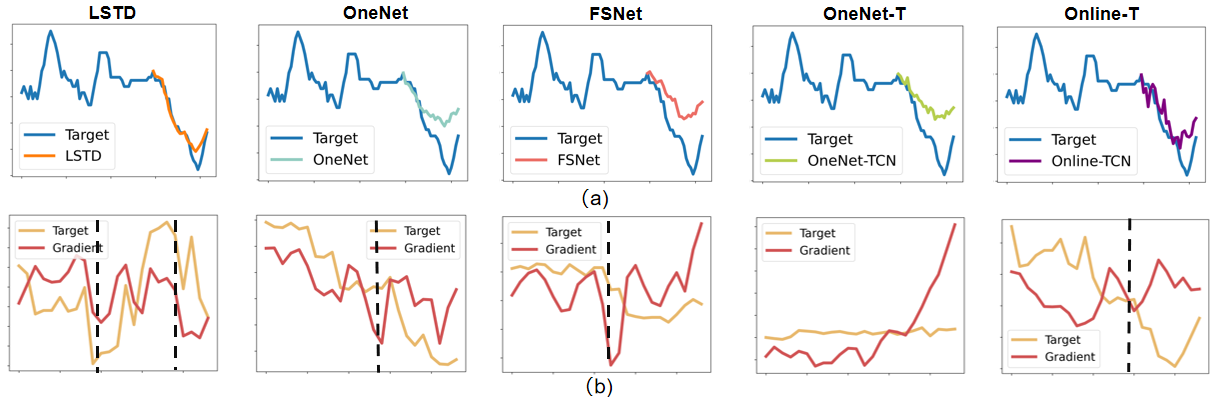}
    \caption{The figure (a) represents the visualization of the proposed LSTD and other baselines. The blue lines denote the ground-truth time series data and the lines with other colors denote the predicted results of different methods. The figure (b) shows the visualization of the LSTD method for detecting interventions. The yellow lines represent the real-time series data, and the red lines represent the gradient.  Black dotted lines denote intervention occurs. \textit{(Best view in color)}
    }
    \label{fig:Visualization}
\end{figure*}

 We further conduct visualization results in the WTH and Exchange dataset in Figure \ref{fig:Visualization}. Remarkably, our method detects interventions well and achieves better visualization results than that of OneNet and FSNet, which do not explicitly disentangle the short-term and long-term variables. This is because the long/short term variables of these methods might be entangled, hindering the rapid adaptation to the changing environment of the data streams, and finally resulting in suboptimal predictions
 . In the meanwhile, our method disentangles the long/short term variables by sparsity dependency constraint, and can efficiently adapt to the new environment. At the same time, the smooth constraint further maintains the long-term variables behind the time series data. Therefore, the prediction curve of our method can well align with the ground truth even if the prediction length is long.
 
 
\begin{table*}[]

\centering
\renewcommand{\arraystretch}{0.65}
\setlength{\tabcolsep}{4pt}
\begin{tabular}{c|c|cccccccccc}
\toprule
Models                     & Len & LSTD                                  & OneNet         & FSNet & OneNet-T & DER++ & ER    & MIR   & TFCL  & Online-T & Informer \\ \midrule
                           & 1   & \textbf{0.347}                        & 0.348          & 0.368 & 0.374    & 0.375 & 0.376 & 0.410 & 0.472 & 0.436    & 0.850    \\
                           & 24  & 0.411                                 & \textbf{0.407} & 0.467 & 0.511    & 0.540 & 0.543 & 0.541 & 0.548 & 0.547    & 0.668    \\
\multirow{-3}{*}{ETTh2}    & 48  & \textbf{0.423}                        & 0.436          & 0.515 & 0.543    & 0.577 & 0.571 & 0.565 & 0.592 & 0.589    & 0.752    \\ \midrule
                           & 1   & \textbf{0.187}                                 & \textbf{0.187} & 0.191 & 0.191    & 0.192 & 0.197 & 0.197 & 0.198 & 0.085    & 0.512    \\
                           & 24  & \textbf{0.217}                        & 0.225          & 0.249 & 0.319    & 0.326 & 0.333 & 0.325 & 0.341 & 0.381    & 0.525    \\
\multirow{-3}{*}{ETTm1}    & 48  & 0.249                                 & \textbf{0.238} & 0.263 & 0.371    & 0.340 & 0.351 & 0.342 & 0.363 & 0.403    & 0.460    \\ \midrule
                           & 1   & \textbf{0.200}                        & 0.201          & 0.216 & 0.221    & 0.235 & 0.244 & 0.244 & 0.240 & 0.276    & 0.458    \\
                           & 24  & \textbf{0.223}                        & 0.225          & 0.276 & 0.345    & 0.351 & 0.356 & 0.355 & 0.363 & 0.367    & 0.417    \\
\multirow{-3}{*}{WTH}      & 48  & \textbf{0.242}                        & 0.279          & 0.301 & 0.356    & 0.359 & 0.363 & 0.361 & 0.382 & 0.362    & 0.419    \\ \midrule
                           & 1   & \textbf{0.226}                        & 0.254          & 0.472 & 0.411    & 0.421 & 0.506 & 0.504 & 0.524 & 0.635    & 0.549    \\
                           & 24  & \textbf{0.292}                        & 0.333          & 0.997 & 0.513    & 1.035 & 1.057 & 1.066 & 1.256 & 1.196    & 1.198    \\
\multirow{-3}{*}{ECL}      & 48  & \textbf{0.294}                        & 0.348          & 1.061 & 0.534    & 1.048 & 1.074 & 1.079 & 1.303 & 1.235    & 1.164    \\ \midrule
                           & 1   & \textbf{0.225}                        & 0.240          & 0.278 & 0.236    & 0.251 & 0.256 & 0.284 & 0.297 & 0.284    & 0.258    \\
                           & 24  & \textbf{0.316}                        & 0.346          & 0.365 & 0.346    & 0.409 & 0.417 & 0.443 & 0.493 & 0.385    & 0.365    \\
\multirow{-3}{*}{Traffic}  & 48  & \textbf{0.332}                        & 0.371          & 0.378 & 0.355    & 0.386 & 0.294 & 0.397 & 0.531 & 0.380    & 0.394    \\ \midrule
                           & 1   & {\color[HTML]{212121} \textbf{0.070}} & 0.085          & 0.174 & 0.117    & 0.173 & 0.124 & 0.118 & 0.153 & 0.169    & 0.115    \\
                           & 24  & \textbf{0.132}                        & 0.148          & 0.206 & 0.166    & 0.227 & 0.210 & 0.204 & 0.227 & 0.213    & 0.196    \\
\multirow{-3}{*}{Exchange} & 48  & \textbf{0.142}                        & 0.170          & 0.254 & 0.173    & 0.243 & 0.241 & 0.209 & 0.183 & 0.258    & 0.217    \\ \bottomrule
\end{tabular}%
\caption{Mean Absolute Error (MAE) results on the different datasets. TCN is abbreviated as T}

\label{MAE}
\end{table*}

\subsection{Ablation Study}
We further devise three model variants. a) \textbf{LSTD}-L1: we remove the interrupted dependency constraint for
short-term disentanglement. b) \textbf{LSTD}-L2: we remove the smooth constraint for long-term
disentanglement. c) \textbf{LSTD}-KL: we remove the long/short-term prior and the corresponding Kullback-Leibler divergence term. Experiment results on the Exchange dataset are shown in Figure \ref{fig:ablation}. We find that 1) the performance of \textbf{LSTD}-L1 drops without an accurate forgetting of the information, implying that the accurate forgetting benefits the quickly adapting to changes in the data domain and improves the disentanglement and forecasting performance. 2) the performance of \textbf{LSTD}-L2 drops without retention of the information, implying that the retention benefits the preserving of the long-term effects and improves the forecasting performance. 3) Both long-term and short-term priors play an important role in forecasting, implying that these priors can capture temporal information.

\section{Summary}
This paper presents a long/short-term state disentanglement model to address the challenges of online time-series forecasting in the presence of nonstationarity led by unknown interventions. Unlike existing methods, this model can theoretically identify both long-term and short-term latent variables, enhancing its relevance to real-world data. Technologically, the LSTD model employs the smooth constraint and sparse dependency constraint to enforce the disentanglement of long/short-term variables. 
In summary, this paper offers valuable insights into enhancing online time-series forecasting via causal representation learning.

\section{Acknowledgments}

This research was supported in part by National Science and Technology Major Project (2021ZD0111501), National Science Fund for Excellent Young Scholars (62122022) and Natural Science Foundation of China (U24A20233, 62206064, 62206061).


\bibliography{aaai24}

 \appendix
 \onecolumn
 \input{supplementary2}



\end{document}


\section{Supplement}  
\newcommand{\beginsupplement}
\ The generating process is defined as follows:
\begin{align}
      [\textbf{x}_{t-\tau-1},\textbf{x}_{t-\tau+1:t},\textbf{x}_{t-\tau}]&=g(\textbf{Z}^d_{t-\tau-1},\textbf{Z}^s_{t-\tau-1:t},\textbf{Z}^d_{t-\tau+1:t},\textbf{Z}^d_{t-\tau})\\
    \textbf{x}_{t-\tau-1}&=g_1(\textbf{Z}^d_{t-\tau-1},\textbf{Z}^s_{t-\tau-1:t},)\\
    \textbf{x}_{t-\tau+1:t},\textbf{x}_{t-\tau}&=g_2(\textbf{Z}^s_{t-\tau-1:t},\textbf{Z}^d_{t-\tau+1:t},\textbf{Z}^d_{t-\tau}
\end{align}
where $\textbf{Z}^s_{t-\tau-1:t} \in \textbf{Z}^s \subset R^{d_{Z_s}},\textbf{Z}^d_{t-\tau-1} \in \textbf{Z}^{d}_1 \subset R^{d_{Z^{d}_1}},\textbf{Z}^d_{t-\tau+1:t},\textbf{Z}^d_{t-\tau} \in \textbf{Z}^d_2 \subset R^{d_{Z^d_2}}$ Both
$g_1$ and $g_2$  are smooth and have non-singular Jacobian matrices
almost anywhere, and g is invertible.
if $\hat{g}_1: Z \longrightarrow V_1 \in \textbf{x}_{t-\tau-1}$ and $\hat{g}_2: Z \longrightarrow V_2 \in \textbf{x}_{t-\tau:t}\cup \textbf{x}$ assume the generating
process of the true model $(g_1, g_2)$ and match the joint distribution $p_{v_1,v_2}$ , then there is a one-to-one mapping between the estimate $\hat{\textbf{Z}}^s_{t-\tau-1:t}$ and the ground truth $\textbf{Z}^s_{t-\tau-1:t}$ over $\textbf{Z}^s \times \textbf{Z}^d \times \textbf{Z}^d $,that is, $\textbf{Z}^s_{t-\tau-1:t}$ is  block-identifiable.

For $(v_1,v_) \sim p_{v_1,v_2}$ 
 because of the matched joint distribution, we have the following relations between the true variables $(\textbf{Z}^s_{t-\tau-1:t},\textbf{Z}^d_{t-\tau-1}, \textbf{Z}^d_{t-\tau+1:t})$ and the estimated ones $(\hat{\textbf{Z}}^s_{t-\tau-1:t},\hat{\textbf{Z}}^d_{t-\tau-1}, \hat{\textbf{Z}}^d_{t-\tau+1:t})$
 \begin{align}
   \textbf{x}_{t-\tau-1}=g_1(\textbf{Z}^s_{t-\tau-1:t},\textbf{Z}^d_{t-\tau-1})
   =\hat{g}_1(\hat{\textbf{Z}}^s_{t-\tau-1:t},\hat{\textbf{Z}}^d_{t-\tau-1})\\
   \textbf{x}_{t-\tau+1:t}=g_2(\textbf{Z}^s_{t-\tau-1:t},\textbf{Z}^d_{t-\tau+1:t})=\hat{g}_2(\hat{\textbf{Z}}^s_{t-\tau-1:t},\hat{\textbf{Z}}^d_{t-\tau+1:t})\\
   (\hat{\textbf{Z}}^s_{t-\tau-1:t},\hat{\textbf{Z}}^d_{t-\tau-1},\hat{\textbf{Z}}^d_{t-\tau+1:t})=\hat{g}^{-1}( \textbf{x}_{t-\tau-1},\textbf{x}_{t-\tau+1:t})\\
   \nonumber=\hat{g}^{-1}(g(\textbf{Z}^s_{t-\tau-1:t},\textbf{Z}^d_{t-\tau-1}, \textbf{Z}^d_{t-\tau+1:t})) := h(\textbf{Z}^s_{t-\tau-1:t},\textbf{Z}^d_{t-\tau-1}, \textbf{Z}^d_{t-\tau+1:t})
 \end{align}
 where we define the smooth and invertible function $h := \hat{g}^{-1} \circ g$ that transforms the true variables $(\textbf{Z}^s_{t-\tau-1:t},\textbf{Z}^d_{t-\tau-1}, \textbf{Z}^d_{t-\tau+1:t})$ to estimates $(\hat{\textbf{Z}}^s_{t-\tau-1:t},\hat{\textbf{Z}}^d_{t-\tau-1}, \hat{\textbf{Z}}^d_{t-\tau+1:t})$
 \begin{align}
     g_1(\textbf{Z}^s_{t-\tau-1:t},\textbf{Z}^d_{t-\tau-1})=\hat{g}_1(h_{\textbf{Z}^s_{t-\tau-1:t},\textbf{Z}^d_{t-\tau-1}}(\textbf{Z}^s_{t-\tau-1:t},\textbf{Z}^d_{t-\tau-1}, \textbf{Z}^d_{t-\tau+1:t}))
 \end{align}
 For $i \in \{1,...,d_{Z^d_1}\}$ and $j \in \{1,...,d_{Z^d_{2}}\}$
taking partial
derivative of the i-th dimension of both sides $w.r.t. z^d_{t-\tau+1:t,j}$
\begin{align}
    0= \frac{\partial \hat{g}_{1,i}(\textbf{Z}^s_{t-\tau-1:t},\textbf{Z}^d_{t-\tau-1})}{\partial z^d_{t-\tau+1:t,j} }=\frac{\partial g_{1,i}(h_{\textbf{Z}^s_{t-\tau-1:t},\textbf{Z}^d_{t-\tau-1}}(\textbf{Z}^s_{t-\tau-1:t},\textbf{Z}^d_{t-\tau-1}))}{\partial z^d_{t-\tau+1:t,j} }
\end{align}
The equation equals zero because there is no $z^d_{t\tau+1:t,j}$ in the left-hand
side of the equation. Expanding the derivative on the right-hand
side gives:
\begin{align}
    \sum_{k \in \{1,...,d_{z^s}+d_{\textbf{Z}^d_{t-\tau-1}}\}} \frac{\partial \hat{g}_{1,i}  }{\partial h_{(\textbf{Z}^s_{t-\tau-1:t},\textbf{Z}^d_{t-\tau-1}),k} } \cdot \frac{\partial  h_{(\textbf{Z}^s_{t-\tau-1:t},\textbf{Z}^d_{t-\tau-1}),k}}{ z^d_{t-\tau+1:t,j} } (\textbf{Z}^s_{t-\tau-1:t},\textbf{Z}^d_{t-\tau-1}, \textbf{Z}^d_{t-\tau+1:t}) = 0 
\end{align}

assume $vectors [\frac{\partial \hat{g}_{1,i}  }{\partial h_{(\textbf{Z}^s_{t-\tau-1:t},\textbf{Z}^d_{t-\tau-1}),1} },..., \frac{\partial \hat{g}_{1,i}  }{\partial
h_{(\textbf{Z}^s_{t-\tau-1:t},\textbf{Z}^d_{t-\tau-1}),d_{z^s}+z^d_{t-\tau-1}}}]$
are linearly
independent, which is equivalent to the non-singular Jacobian matrix condition. Therefore, the linear system
is invertible and the solution states that:
\begin{align}
    \frac{\partial  h_{(\textbf{Z}^s_{t-\tau-1:t},\textbf{Z}^d_{t-\tau-1}),k}}{ z^d_{t-\tau+1:t,j} }(\textbf{Z}^s_{t-\tau-1:t},\textbf{Z}^d_{t-\tau-1}, \textbf{Z}^d_{t-\tau+1:t}) = 0
\end{align}

 Therefore, we have shown that $h_{\textbf{Z}^s_{t-\tau-1:t},\textbf{Z}^d_{t-\tau-1}} , i.e. (\hat{\textbf{Z}}^s_{t-\tau-1:t},\hat{\textbf{Z}}^d_{t-\tau-1})$ does
not depend on $\textbf{Z}^d_{t-\tau+1:t}$ 

Applying the same reasoning to $h_{\textbf{Z}^s_{t-\tau-1:t},\textbf{Z}^d_{t-\tau+1}} $ , we can obtain \\ that $h_{\textbf{Z}^s_{t-\tau-1:t},\textbf{Z}^d_{t-\tau+1}} , i.e. (\hat{\textbf{Z}}^s_{t-\tau-1:t},\hat{\textbf{Z}}^d_{t-\tau+1})$  does not depend on $\textbf{Z}^d_{t-\tau-1}$
Thus,for $(\hat{\textbf{Z}}^s_{t-\tau-1:t},\hat{\textbf{Z}}^d_{t-\tau-1}, \hat{\textbf{Z}}^d_{t-\tau+1:t})$ we can observe that $\hat{\textbf{Z}}^s_{t-\tau-1:t}$ does not depend on  $\textbf{Z}^d_{t-\tau+1:t}$ and $\textbf{Z}^d_{t-\tau-1}$ , that is , $\hat{\textbf{Z}^s_{t-\tau-1:t}} = h_{\textbf{Z}^s_{t-\tau:t}}(\textbf{Z}^s_{t-\tau-1:t})$
$\textbf{Z}^s_{t-\tau-1:t}$ are subspace identifiable. \\

assumption
\begin{itemize}
    \item We imposed an L1 constraint on $z^d$, so the data processiong is identifiable\\
    \item Assuming intervention occurs within at least $\tau >2$ steps\\
    \item  (Smooth and Positive Density:) The probability density function of latent variables is smooth and positive, i.e.,$p(\textbf{z}_{t-\tau+1:t}|\textbf{z}^d_{t-\tau})>0$ over $Z_{t-\tau+1:t}$\\
    \item (Conditional independent:) Conditioned on $\textbf{z}^d_{t-\tau}$ each $z_i$ is independent of any other $z_j$ for $i,j\in {1,...,n}, i\neq j,i.e$, $\log{p(\textbf{z}^d_{t-\tau+1:t}|\textbf{z}^d_{t-\tau})} = \sum_{k=1}^{n_s} \log {p(z^d_{t-\tau+1:t,k}|\textbf{z}^d_{t-\tau})}$ \\
    \item (Linear Independent:) For any $\textbf{z}^d\in Z^d_{t-\tau+1:t}\subseteq \mathbb{R}^{n_s},\bar{v}_{t-\tau,1},...,\bar{v}_{t-\tau,n_s}$ as $n_s$ vector functions in $z^d_{t-\tau,1},...,z^d_{t-\tau.l},...,z^d_{t-\tau,n_s}$ are linear independent, where $\bar{v}_{t-\tau,l}$ are formalized as follows:
    \begin{align}
        \bar{\textbf{v}}_{t-\tau,l}=\frac{\partial^2 \log{p(\textbf{z}^d_{t-\tau+1:t}|\textbf{z}^d_{t-\tau})}  }{\partial z^d_{t-\tau+1:t ,k} \partial z^d_{t-\tau,k} } 
    \end{align}

\end{itemize}
 Proof. We start from the matched marginal distribution to develop the relation between $\textbf{z}_{t-\tau+1:t}$ and $\hat{\textbf{z}}_{t-\tau+1:t}$ as follows:

\begin{align}  
    p(\hat{\textbf{z}}_{t-\tau+1:t}) = p(\textbf{z}_{t-\tau+1:t})  \Longleftrightarrow p(\hat{g}(\hat{\textbf{z}}_{t-\tau+1:t})) = p(g(\textbf{z}_{t-\tau+1:t}))
    \Longleftrightarrow \\
    \nonumber
    p(g^{-1}\circ \hat{g}(\hat{\textbf{z}}_{t-\tau+1:t}))
    = p(\textbf{z}_{t-\tau+1:t})|\textbf{J}_{g^{-1}}|\Longleftrightarrow p(h(\hat{\textbf{z}}_{t-\tau+1:t})) = p(\textbf{z}_{t-\tau+1:t}),
\end{align}

\begin{gather}  
    p(\hat{\textbf{z}}_{t-\tau+1:t}|\textbf{z}^d_{t-\tau}) = p(\textbf{z}_{t-\tau+1:t}|\textbf{z}^d_{t-\tau}) \cdot |\textbf{J}_h| \\  
    \log{p(\hat{\textbf{z}}_{t-\tau+1:t}|\textbf{z}^d_{t-\tau})} = \log{p(\textbf{z}_{t-\tau+1:t}|\textbf{z}^d_{t-\tau})} \\+ \log{|\textbf{J}_h|}   
    = \log{p(\textbf{z}^d_{t-\tau+1:t}|\textbf{z}^d_{t-\tau})} + \log{p(\textbf{z}^s_{t-\tau+1:t}|\textbf{z}^d_{t-\tau})} + \log{|\textbf{J}_h|}  
\end{gather}
Therefore, for $i \in \{n_s+1,...,n\}$,the partial derivative of the equation $w.r.t\ \hat{z}^s_{t-\tau+1:t,i}$ is:

\begin{align}
    \frac{\partial p(\hat{z}_{t-\tau+1:t}|\textbf{z}^d_{t-\tau}) }{\partial \hat{z}^s_{t-\tau+1:t,i} } =  {\textstyle \sum_{k=1}^{n_s}}\frac{\partial \log{p(z^s_{t-\tau+1:t,k}|\textbf{z}^d_{t-\tau})} }{\partial z^s_{t-\tau+1:t,k} } \cdot  \frac{\partial z^s_{t-\tau+1:t,k} }{\partial \hat{z}^s_{t-\tau+1:t,i}  } \\ +
     {\textstyle \sum_{k=n_s+1}^{n}} \frac{\partial p(z^d_{t-\tau+1:t}|\textbf{z}^s_{t-\tau})}{\partial z^d_{t-\tau+1:t,k} } \cdot \frac{\partial z^d_{t-\tau+1:t,k} }{\partial \hat{z}^s_{t-\tau+1:t,i}  } + \frac{\partial \log{|J_h|} }{\hat{z}^s_{t-\tau+1:t,i}} 
\end{align}

Sequentially, for each $l\in \{1,...,n_s\}$ , and each value of $z^d_{t-\tau,l}$ , its partial derivative $w.r.t.\ z^d_{t-\tau,l}$  is shown as follows:

\begin{align}
    \frac{\partial^2 p(\hat{z}_{t-\tau+1:t}|\textbf{z}^d_{t-\tau}) }{\partial \hat{z}^s_{t-\tau+1:t,i} \partial z^d_{t-\tau,l}} =  {\textstyle \sum_{k=1}^{n_s}}\frac{\partial^2 \log{p(z^s_{t-\tau+1:t,k}|\textbf{z}^d_{t-\tau})} }{\partial z^s_{t-\tau+1:t,k} \partial z^d_{t-\tau,l}} \cdot  \frac{\partial z^s_{t-\tau+1:t,k} }{\partial \hat{z}^s_{t-\tau+1:t,i}  } +
     \\
     \nonumber{\textstyle \sum_{k=n_s+1}^{n}} \frac{\partial^2 p(z^d_{t-\tau+1:t}|\textbf{z}^d_{t-\tau})}{\partial z^d_{t-\tau+1:t,k} \partial z^d_{t-\tau,l}} \cdot \frac{\partial z^d_{t-\tau+1:t,k} }{\partial \hat{z}^s_{t-\tau+1:t,i}  } + \frac{\partial^2 \log{|J_h|} }{\partial z^d_{t-\tau:t,i}\partial \hat{z}^s_{t-\tau+1,l}} 
\end{align}

\begin{align}
    0 =  {\textstyle \sum_{k=1}^{n_s}}\frac{\partial^2 \log{p(z^d_{t-\tau+1:t,k}|\textbf{z}^d_{t-\tau})} }{\partial z^d_{t-\tau+1:t,k} \partial z^d_{t-\tau,l}} \cdot  \frac{\partial z^d_{t-\tau+1:t,k} }{\partial \hat{z}^s_{t-\tau+1:t,i}  }
\end{align}

Identifiablity under a Fixed Temporal Causal Model\\
\begin{itemize}
    \item (Conditional independent:) Conditioned on $x_{t-1}$ each $z_{t,i}$ is independent of any other $z_j$ for $i,j\in {1,...,n}, i.e$, $\log{p(\textbf{z}^s_{t}|\textbf{z}_{t-1})} = \sum_{k=1}^{n_s} \log {p(z^s_{t,k}|\textbf{z}_{t-1})}$ \\
    \item \begin{align}
    v_{k,t} = (\frac{\partial^2 \log{p(z^s_{t,k}|z^s_{t-1})} }{\partial z^s_{k,t} \partial z^s_{1,t-1}},\frac{\partial^2 \log{p(z^s_{t,k}|z^s_{\tau -1}})} {\partial z^s_{k,t} \partial z^s_{2,t-1}},...,\frac{\partial^2 \log{p(z^s_{t,k}|z^s_{t-1}})} {\partial z^s_{k,t} \partial z^s_{n,t-1}} )^T\\
    \nonumber
    \hat{v}_{k,t} = (\frac{\partial^3 \log{p(z^s_{t,k}|z^s_{t-1})} }{\partial^2 z^s_{k,t} \partial z^s_{1,t-1}}, \frac{\partial^3 \log{p(z^s_{t,k}|z^s_{t-1})} }{\partial^2 z^s_{k,t} \partial z^s_{2,t-1}},..., \frac{\partial^3 \log{p(z^s_{t,k}|z^s_{t-1})} }{\partial^2 z^s_{k,t} \partial z^s_{n,t-1}})^T
\end{align}
\item \begin{align}
    z^s_t = h(\hat{z}^s_t)
\end{align}
\end{itemize}

Since $p(\hat{z}^s_t,\hat{z}^s_{t-1}) =p(\hat{z}^s_t|\hat{z}^s_{t-1})p(\hat{z}^s_{t})$ \\while $p(\hat{z}^s_{t})$ does not involve $z^s_{t,i}$ or $z^s_{t,j}$ the above equality is equivalent to:

\begin{align}
    \frac{\partial^2 \log{p(\hat{z}^s_t|\hat{z}^s_{t-1})} }{\partial z^s_{t,i}\partial z^s_{t,j} } =0
\end{align}

\begin{align}
    \log{p(\hat{z}^s_t|\hat{z}^s_{t-1})} & = \log{p(z^s_t|z^s_{t-1})} + \log{|J_{h}|}\\
    \nonumber
   & =  \textstyle \sum_{k=1}^{n} \log{p(z^s_{t,k}|z^s_{t-1})} + \log{|J_{h}|}
\end{align}

Its partial derivative $w.r.t. \hat{z}^s_{t,i}$

\begin{align}
    \frac{\partial \log{p(\hat{z}^s_t|\hat{z}^s_{t-1}) }}{\partial \hat{z}^s_{t,i} } =  \textstyle \sum_{k=1}^{n} \frac{\partial p(\hat{z}^s_t|\hat{z}^s_{t-1})}{\partial z^s_{t,k}}\frac{\partial z^s_{t,k}}{\partial \hat{z}^s_{t,i}} - \frac{\partial \log{|J_h|} }{\partial \hat{z}^s_{t,i}} 
\end{align}

Its second-order cross derivative is:

\begin{align}
    \frac{\partial^2 \log{p(\hat{z}^s_t|\hat{z}^s_{t-1}) }}{\partial \hat{z}^s_{t,i} \partial \hat{z}^s_{t,j}} &=  \textstyle \sum_{k=1}^{n} (\frac{\partial^2 p(\hat{z}^s_t|\hat{z}^s_{t-1})}{\partial z^{s2}_{t,k}}\frac{\partial z^s_{t,k}}{\partial \hat{z}^s_{t,i}}\frac{\partial z^s_{t,k}}{\partial \hat{z}^s_{t,j}} + \frac{\partial p(\hat{z}^s_t|\hat{z}^s_{t-1})}{\partial z^s_{t,k}} \cdot \frac{\partial^2 z^s_{t,k}}{\partial \hat{z}^s_{t,i}  \partial \hat{z}^s_{t,j}})\\
    \nonumber &- \frac{\partial \log{|J_h|} }{\partial\hat{z}^s_{t,i}\partial \hat{z}^s_{t,j}} 
\end{align}

The above quantity is always 0, Therefore, for each$l = 1, 2, ..., n$ and each value $z^s_{t-1,l}, $ its partial derivative$w.r.t. z^s_{t-1,l} $ is always 0. That is,

\begin{align}
    \frac{\partial^3 \log{p(\hat{z}^s_t|\hat{z}^s_{t-1}) }}{\partial \hat{z}^s_{t,i} \partial \hat{z}^s_{t,j} \partial z^s_{t-1,l}} &= \textstyle \sum_{k=1}^{n} (\frac{\partial^3
p(\hat{z}^s_t|\hat{z}^s_{t-1})}{\partial z^{s2}_{t,k}\partial z^s_{t-1,l}}
\frac{\partial z^s_{t,k}}{\partial \hat{z}^s_{t,i}}
\frac{\partial z^s_{t,k}}{\partial \hat{z}^s_{t,j}} \\
\nonumber &+ 
\frac{\partial^2 p(\hat{z}^s_t|\hat{z}^s_{t-1})}{\partial z^s_{t,k}\partial z^s_{t-1,l}}
\frac{\partial^2 z^s_{t,k}}{\partial \hat{z}^s_{t,i} \cdots 
\partial \hat{z}^s_{t,j}}) \equiv  0
\end{align}

%% file: math_commands.tex
\usepackage{amsmath,amsfonts,bm}

\def\eqref#1{equation~\ref{#1}}

\def\1{\bm{1}}

\def\rvx{{\mathbf{x}}}

\def\rvz{{\mathbf{z}}}

\DeclareMathAlphabet{\mathsfit}{\encodingdefault}{\sfdefault}{m}{sl}
\SetMathAlphabet{\mathsfit}{bold}{\encodingdefault}{\sfdefault}{bx}{n}



%% file: supplementary2.tex
\renewcommand{\theequation}{A.\arabic{equation}}
\section{A. Related Works}\label{app:related_works}
\subsection{Time Series Forecasting}
Recently, various research studies focused on time series forecasting problems, and deep learning-based methods have been very successful in this field. More precisely, the deep learning-based methods can be divided into several classes. First, the model based on RNN utilizes a recursive structure with memory to construct hidden layer transitions over time points .\cite{graves2012long,lai2018modeling,salinas2020deepar}. Second, TCN \cite{bai2018empirical,wang2023micn,wu2022timesnet} based approach to modeling hierarchical temporal patterns and extracting features using a shared convolutional kernel. Besides, there are simple but very effective methods as well, such as based on MLP \cite{oreshkin2019n,zeng2023transformers,zhang2022less,li2024and} and on states-base-model \cite{gu2022parameterization,gu2021combining,gu2021efficiently}. 
Above these methods, the Transformer methods are especially outstanding and get the great process on the time series forecasting task \cite{kitaev2020reformer,liu2021pyraformer,wu2021autoformer,zhou2021informer}.
However, these existing methods are based on offline data processing, contrary to the mainstream ONLINE training methods of the significant data era. Since the above methods are unsuitable for direct application to online problems, a model that can be trained on online data and perform well is needed.
\subsection{Online Time Series Forecasting}
Due to the rapid increase in train data and the requirement for model updates online, online time series forecasting has become more popular than offline ones\cite{anava2013online,liu2016online,gultekin2018online,aydore2019dynamic}.
\cite{pan2024structural} employs structural consistency regularization to capture a range of scenarios, using a representation-matching memory replay scheme to retain temporal dynamics and dependencies. \cite{luan2024online} applies tensor factorization for streaming tensor time series prediction, updating the predictor in a low-complexity online manner to adapt to evolving data. \cite{mejri2024novel} addresses online nonlinear time-series forecasting by mapping low-dimensional series to high-dimensional spaces for linear hyperdimensional prediction, adapting to temporal distribution shifts.Online time series forecasting is a widely used technique in the real world due to the continuity of the data and the frequent drift of concepts. In this approach, the learning process occurs over a series of rounds. The model receives a look-back window, predicts the forecasting window, and then displays the valid values to improve the model's performance in the next round. Recently, a brunch of online time series forecasting work got excellent results, including considering gradient updates to optimize fast new as well as retained information \cite{pham2022learning} and models that consider both temporal and feature dimensions \cite{wen2024onenet}. Nevertheless, the fast adaptation and information retention of the models mentioned above are simultaneous. It needs to decouple the long and short term, which can lead to confounding results and suboptimal predictions. To solve this problem, the LSTD decouples the data first to isolate the long and short-term effects on the prediction, with the long-term effects being used to preserve the characteristics of the historical data and the short-term effects being used to quickly adapt to changes in the data for better online prediction.
\subsection{Continual Learning}
Continual learning is a novel topic and aims to set up intelligent agency by learning the sequence of tasks to perform with restricted access to experience \cite{lopez2017gradient}. A continual learning model must balance the knowledge of the current task and the prompt of the future learning process, as known in the stability-plasticity dilemma \cite{lin1992self,grossberg2013adaptive}. Due to their connection to how humans learn, several neuroscience frameworks have prompted the development of various continual learning algorithms. The continual learning model corresponds to the requirement of online time series forecasting. The constant learning can enable real-time updates upon receiving the new data to adapt the data dynamics better, improving the model accuracy.
The proposed LSTD incorporates continual learning into an online time series forecasting model, which mitigates the stability-plasticity problem by decoupling the long and short-term effects, retaining the knowledge of previous tasks through the long-term effects, and facilitating the learning of future tasks through the short-term effects.
\subsection{Causal Representation Learning}
To recover the latent variable with identification guarantees\cite{yao2023multi,scholkopf2021toward,liu2023causal,gresele2020incomplete}, independent component analysis (ICA) has been used in a number of studies to determine the casual representation \cite{rajendran2024learning,mansouri2023object,wendong2024causal,li2024identification}. 
Conventional approaches presuppose a linear mixing function for latent and observable variables. \cite{comon1994independent,hyvarinen2013independent,lee1998independent,zhang2007kernel}. However, determining the linear mixing function is a difficult problem in real-world situations. For the identifiability, many assumptions are made throughout the nonlinear ICA process, including the sparse generation process and the usage of auxiliary variables\cite{zheng2022identifiability,hyvarinen1999nonlinear,hyvarinen2024identifiability,khemakhem2020ice,li2023identifying}.\\
Specifically, Aapo et al.'s study confirms identifiability first. The exponential family is assumed to consist of latent sources in Ref. \cite{khemakhem2020variational,hyvarinen2016unsupervised,hyvarinen2017nonlinear, hyvarinen2019nonlinear}, where auxiliary variables such as domain indexes, time indexes, and class labels are added. Furthermore, Zhang et al.'s study \cite{kong2022partial, xie2023multi,kong2023identification,yan2024counterfactual} demonstrates that the exponential family assumption is not necessary to accomplish component-wise identification for nonlinear ICA. \\
Sparsity assumptions were used in several study endeavors to attain identifiability without supervised signals\cite{zheng2022identifiability,hyvarinen1999nonlinear,hyvarinen2024identifiability,khemakhem2020ice,li2023identifying}. For example, Lachapelle et al. \cite{lachapelle2023synergies, lachapelle2022partial} presented mechanism sparsity regularization as an inductive bias to find causative latent components. Zhang et al. selected the latent variable sparse structures in Ref. \cite{zhang2024causal} to achieve identifiability under distribution shift. Furthermore, nonlinear ICA was utilized in Ref. \cite{hyvarinen2016unsupervised,yan2024counterfactual,huang2023latent,halva2020hidden,lippe2022citris} to get the identifiability of time series data. 
Premise and capitalization of variance variations across data segments based on separate sources were utilized in the study by Aapo et al.\cite{hyvarinen2016unsupervised} to detect nonstationary time series data identifiability. Conversely, the latent variables in stationary time series data are found via permutation-based contrastive learning. Independent noise and variability history information features have been used more recently in TDRL \cite{yao2022temporally} and LEAP \cite{yao2021learning}. \\
Simultaneously, latent variables were discovered by Song et al. \cite{song2024temporally} without the need to observe the domain variables. Imant et al. \cite{daunhawer2023identifiability} described multimodal comparative learning as having identifiability in terms of modality. Yao et al.\cite{yao2023multi} postulated that multi-perspective causal representations can still be identified when there are incomplete observations. This paper uses multi-modality time series data and leverages historical variability information and multi-modality data fairness to demonstrate identifiability.

\renewcommand{\theequation}{B.\arabic{equation}}
\section{B. Proof}

\subsection{Identification}
In this section, we provide the definition of different types of identification.
\subsubsection{Subspace Identification}
For each ground-truth changing latent variables $z_{t,i}$, the subspace identification means that there exists $\hat{\rvz}_{t}$ and an invertible function $z_{t,i}=h_i(\hat{\rvz}_t)$, such that $z_{t,i}=h_i(\hat{\rvz}_t)$.


\subsection{Identification Guarantees}

\subsubsection{Subspace of Identification Long/Short-term Latent Variables }
\begin{theorem}
	(\textbf{Subspace of Identification Long/Short-term Latent Variables $\rvz^s_t$ and $\rvz^d_t$.}) We follow the data generation process in Figure 1 (c) and Equation (1)-(3) and make the following assumptions:
	\begin{itemize} 
		\item A1 \underline{(Smooth, Positive and Conditional independent Density:)} \cite{yao2022temporally,yao2021learning} The probability density function of latent variables is smooth and positive, i.e., $p(\rvz_{t-\tau+1:t}|\rvz_{t-\tau})>0$ over $\rvz_{t-\tau}$ and $\rvz_{t-\tau+1:t}$.
		Conditioned on $\rvz_{t-\tau}$ each $z_i$ is independent of any other $z_j$ for $i,j\in {1,...,n}, i\neq j,\ i.e$, $\log{p(\rvz_{t-\tau+1:t}|\rvz_{t-\tau})} = \sum_{k=1}^{n_s} \log {p(z_{t-\tau+1:t,k}|\rvz_{t-\tau})}$
		\item A2 \underline{(non-singular Jacobian):} \cite{kong2023understanding} Each generating function $g$ has non-singular Jacobian matrices almost anywhere and $g$ is invertible. 
		\item A3\underline{ (Linear Independence:) }\cite{yao2022temporally} For any $\rvz^d\in Z^d_{t-\tau+1:t}\subseteq {R}^{n_d},\ \bar{v}_{t-\tau,1},\ ...,\ \bar{v}_{t-\tau,n_d}$ as $n_d$ vector functions in $z^d_{t-\tau,1},\ ...,z^d_{t-\tau.l},\ ...,\ z^d_{t-\tau,n_d}$ are linear independent, where $\bar{v}_{t-\tau,l}$ are formalized as follows:
		\begin{align}
			\bar{\textbf{v}}_{t-\tau,l}=\frac{\partial^2 \log{p(\rvz^d_{t-\tau+1:t}|\rvz^d_{t-\tau})}  }{\partial z^d_{t-\tau+1:t ,k} \partial z^d_{t-\tau,l} } 
		\end{align}
	\end{itemize}
	Suppose that we learn $(\hat{g}, {\hat{f}}_i^s, {\hat{f}}_i^d)$ to achieve Equation (1)-(3) with the minimal number of transition edge among short term latent variables $\rvz^d_1, \cdots, \rvz^d_t, \cdots$, then the long-term and short-term latent variables are block-wise identifiable.
\end{theorem}
\begin{proof}
	First, we can view the data generation process as two parts. Separating the time series is helpful for us to identify, as interventions occur in the time series process, The specific process is as follows:
	\begin{align}
		\textbf{x}_{t-\tau-1:t}&=g(\rvz^d_{t-\tau-1:t},\rvz^s_{t-\tau-1:t})\\
		\textbf{x}_{t-\tau-1}&=g_1(\rvz^d_{t-\tau-1},\rvz^s_{t-\tau-1:t},)\\
		\textbf{x}_{t-\tau+1:t}&=g_2(\rvz^d_{t-\tau+1:t},\rvz^s_{t-\tau-1:t}),
	\end{align}
	where time lag is $\tau+1$,  $\rvz^s_{t-\tau-1:t} \in \rvz^s \subset R^{{n_s}},\ \rvz^d_{t-\tau-1} \in \rvz^{d}_1 \subset R^{{n^1_d}},\ \rvz^d_{t-\tau+1:t} \in \rvz^d_2 \subset R^{n^2_d}$. 
	This involves an intervention at time $t-\tau-1$, resulting in an interrupted edge between short-term latent $\rvz_{t-\tau-1}^d$ and  $\rvz_{t-\tau}^d$. Whereas $\rvz^s_{t-\tau-1:t}$ represents a long-term latent, it is not subject to intervention interruption, and therefore is shared by the two components. Of note, the first $z^d$ after the break is unrecognizable because it lacks supervisory signals; hence, we omit $\rvz^d_{t-\tau}$ here.And We will take $\rvz^d_{t-\tau}$ as the supervision signal for the subsequent consecutive points. Besides, Both
	$g_1$ and $g_2$  are smooth and have non-singular Jacobian matrices
	almost anywhere. Let’s denote $\hat{g}_1: Z \longrightarrow V_1 \in \textbf{x}_{t-\tau-1}$ and $\hat{g}_2: Z \longrightarrow V_2 \in \textbf{x}_{t-\tau+1:t}$, the generating
	process of the true model $(g_1, g_2)$ and match the joint distribution $p_{v_1,v_2}$

	For $(v_1,v_2) \sim p_{v_1,v_2}$ 
	because of the matched joint distribution, we have the following relations between the true variables $(\rvz^s_{t-\tau-1:t},\rvz^d_{t-\tau-1}, \rvz^d_{t-\tau+1:t})$ and the estimated ones $(\hat{\rvz}^s_{t-\tau-1:t},\hat{\rvz}^d_{t-\tau-1}, \hat{\rvz}^d_{t-\tau+1:t})$:
	\begin{align}
		\label{the:1}
		\textbf{x}_{t-\tau-1}&=g_1(\rvz^s_{t-\tau-1:t},\rvz^d_{t-\tau-1})
		=\hat{g}_1(\hat{\rvz}^s_{t-\tau-1:t},\hat{\rvz}^d_{t-\tau-1})
	\end{align}
	\begin{align}
		\label{the:2}
		\textbf{x}_{t-\tau+1:t}&=g_2(\rvz^s_{t-\tau-1:t},\rvz^d_{t-\tau+1:t})=\hat{g}_2(\hat{\rvz}^s_{t-\tau-1:t},\hat{\rvz}^d_{t-\tau+1:t})
	\end{align}
	\begin{align}
		\label{the:3}
		(\hat{\rvz}^s_{t-\tau-1:t},\hat{\rvz}^d_{t-\tau-1},\hat{\rvz}^d_{t-\tau+1:t})&=\hat{g}^{-1}( \textbf{x}_{t-\tau-1},\textbf{x}_{t-\tau+1:t})
		=\hat{g}^{-1}(g(\rvz^s_{t-\tau-1:t},\rvz^d_{t-\tau-1}, \rvz^d_{t-\tau+1:t}))\\ \nonumber &:= h(\rvz^s_{t-\tau-1:t},\rvz^d_{t-\tau-1}, \rvz^d_{t-\tau+1:t}),
	\end{align}
	where $\hat{g}_1, \hat{g}_2$ are the estimated invertible generating function and we define the smooth and invertible function $h:= \hat{g}^{-1} \circ g$ that transforms the true variables $(\rvz^s_{t-\tau-1:t},\rvz^d_{t-\tau-1}, \rvz^d_{t-\tau+1:t})$ to estimates $(\hat{\rvz}^s_{t-\tau-1:t},\hat{\rvz}^d_{t-\tau-1}, \hat{\rvz}^d_{t-\tau+1:t})$
	By combining Equation (\ref{the:3}) and (\ref{the:1}), we have :
	\begin{align}
		\label{the:4}
		g_1(\rvz^s_{t-\tau-1:t},\rvz^d_{t-\tau-1})=\hat{g}_1(h_{\rvz^s_{t-\tau-1:t},\rvz^d_{t-\tau-1}}(\rvz^s_{t-\tau-1:t},\rvz^d_{t-\tau-1}, \rvz^d_{t-\tau+1:t})),
	\end{align}
	For $i \in \{1,...,n^1_d\}$ and $j \in \{1,...,n^2_d\}$
	taking partial
	derivative of the j-th dimension of both sides of Equation (\ref{the:4}) $w.r.t.\  z^d_{t-\tau+1:t,j}$ and have:
	\begin{align}
		0= \frac{\partial g_{1,i}(\rvz^s_{t-\tau-1:t},\rvz^d_{t-\tau-1})}{\partial z^d_{t-\tau+1:t,j} }=\frac{\partial \hat{g}_{1,i}(h_{\rvz^s_{t-\tau-1:t},\rvz^d_{t-\tau-1}}(\rvz^s_{t-\tau-1:t},\rvz^d_{t-\tau-1}))}{\partial z^d_{t-\tau+1:t,j} },
	\end{align}
	The aforementioned equation equals 0 because there is no $z^d_{t-\tau+1:t,j}$ in the left-hand side of the equation. By expanding the derivative on the right-hand side, we further have:
	\begin{align}
		\sum_{k \in \{1,...,n_s+n^1_{d}\} }\frac{\partial \hat{g}_{1,i} }{\partial h_{(\rvz^s_{t-\tau-1:t},\rvz^d_{t-\tau-1}),k} } \cdot \frac{\partial  h_{(\rvz^s_{t-\tau-1:t},\rvz^d_{t-\tau-1}),k}}{ \partial z^d_{t-\tau+1:t,j} } (\rvz^s_{t-\tau-1:t},\rvz^d_{t-\tau-1}, \rvz^d_{t-\tau+1:t}) = 0,
	\end{align}
	Since $\hat{g}_1$ is invertible, the determinant of $\mathrm{J}_{\hat{g}_1}$ does not equal to 0, meaning that for $n_s+n^1_{d}$ different values of $\hat{g}_{1,i}$, each vector $[\frac{\partial \hat{g}_{1,i}  }{\partial h_{(\rvz^s_{t-\tau-1:t},\rvz^d_{t-\tau-1}),1} },..., \frac{\partial \hat{g}_{1,i}  }{\partial
		h_{(\rvz^s_{t-\tau-1:t},\rvz^d_{t-\tau-1}),n_s+n^1_{d}}}]$ are linearly independent. Therefore, the $(n_s+n^1_d)\times (n_s+n^1_d)$ linear system is invertible and has the unique solution as follows:
	\begin{align}
		\label{eg:s2}
		\frac{\partial  h_{(\rvz^s_{t-\tau-1:t},\rvz^d_{t-\tau-1}),k}}{ z^d_{t-\tau+1:t,j} }(\rvz^s_{t-\tau-1:t},\rvz^d_{t-\tau-1}, \rvz^d_{t-\tau+1:t}) = 0,
	\end{align}
	According to Equation (\ref{eg:s2}) for any $k\in \{1,\cdots,n_s+n^1_{d}\}$ and $j\in\{1,\cdots,n^2_{d}\}$,  $h_{(\rvz^s_{t-\tau-1:t},\rvz^d_{t-\tau-1}),k}$ does not depend on $\rvz_{t-\tau+1:t}^{d}$. In other word, $\{\hat{\rvz}^s_{t-\tau-1:t},\rvz^d_{t-\tau-1}\}$ does not depend on $\hat{\rvz}^d_{t-\tau+1:t}$. 
	Applying the same reasoning to $h_{(\rvz^s_{t-\tau-1:t},\rvz^d_{t-\tau+1:t})} $, we can obtain that $h_{(\rvz^s_{t-\tau-1:t},\rvz^d_{t-\tau+1:t})},\  i.e. (\hat{\rvz}^s_{t-\tau-1:t},\hat{\rvz}^d_{t-\tau+1})$  does not depend on $\rvz^d_{t-\tau-1}$. 
	Thus, for $(\hat{\rvz}^s_{t-\tau-1:t},\hat{\rvz}^d_{t-\tau-1}, \hat{\rvz}^d_{t-\tau+1:t})$ we can observe that $\hat{\rvz}^s_{t-\tau-1:t}$ does not depend on  $\rvz^d_{t-\tau+1:t}$ and $\rvz^d_{t-\tau-1}$, that is, $\hat{\rvz}^s_{t-\tau-1:t} = h_{\rvz^s_{t-\tau-1:t}}(\rvz^s_{t-\tau-1:t})$, 
	$\rvz^s_{t-\tau-1:t}$ are subspace identifiable.

	Since the matched marginal distribution of $p(\rvz_{t-\tau+1:t}|\rvx_{t-\tau})$, we have:
	\begin{align}
		\quad p(\hat{\rvx}_{t-\tau+1:t}|\rvx_{t-\tau})=p(\rvx_{t-\tau+1:t}|\rvx_{t-\tau}) \Longleftrightarrow p(\hat{g}(\hat{\rvz}_{t-\tau+1:t})|\rvx_{t-\tau})=p(g(\rvz_{t-\tau+1:t})|\rvx_{t-\tau}),
	\end{align}
	
	where $\rvz_{t-\tau+1:t}=\{\rvz_{t-\tau+1:t}^s,\rvz^d_{t-\tau+1:t}\}$ and $\hat{\rvz}_{t-\tau+1:t}=\{\hat{\rvz}_{t-\tau+1:t}^s,\hat{\rvz}^d_{t-\tau+1:t}\}$.Sequentially, by using the change of variables formula, we can further obtain Equation (\ref{the5})
	\begin{equation}
		\begin{split}
			\label{the5}
			p(\hat{g}(\hat{\rvz}_{t-\tau+1:t})|\rvx_{t-\tau})=p(g(\rvz_{t-\tau+1:t})|\rvx_{t-\tau}) &\Longleftrightarrow p(g^{-1}\circ \hat{g}(\hat{\rvz}_{t-\tau+1:t})|\rvx_{t-\tau})|\mathbf{J}_{g^{-1}}|=p(\rvz_{t-\tau+1:t}|\rvx_{t-\tau})|\mathbf{J}_{g^{-1}}|\\&\Longleftrightarrow p(h(\hat{\rvz}_{t-\tau+1:t})|\rvx_{t-\tau})=p(\rvz_{t-\tau+1:t}|\rvx_{t-\tau})\\&\Longleftrightarrow p(h(\hat{\rvz}_{t-\tau+1:t})|\hat{\rvz}_{t-\tau})=p(\rvz_{t-\tau+1:t}|\rvz_{t-\tau}),
		\end{split}
	\end{equation}
	where $h:=g^{-1}\circ \hat{g}_{1}$ is the transformation between the ground-true and the estimated latent variables. $\mathbf{J}_{g^{-1}}$ denotes the absolute value of Jacobian matrix determinant of $g^{-1}$. Since we assume that $g$ and $\hat{g}$ are invertible, $|\mathbf{J}_{g^{-1}}|\neq 0$ and $h$ is also invertible.

	According to the A1 (conditional independent assumption), we can have Equation (\ref{the6})
	\begin{align}
		\label{the6}
		p(\rvz_{t-\tau+1:t}|\rvz_{t-\tau})=\prod_{i=1}^n p(z_{{t-\tau+1:t},i}|\rvz_{t-\tau}); \\ \nonumber\quad p(\hat{\rvz}_{t-\tau+1:t}|\hat{\rvz}_{t-\tau})=\prod_{i=1}^n p(\hat{z}_{{t-\tau+1:t},i}|\hat{\rvz}_{t-\tau}).
	\end{align}
	For convenience, we take logarithm on both sides of Equation (\ref{the6}) and have:
	\begin{align}
		\label{the7}
		\log p(\rvz_{t-\tau+1:t}|\rvz_{t-\tau}) = \sum_{i=1}^n \log p(z_{{t-\tau+1:t},i}|\rvz_{t-\tau});\\ \nonumber \quad \log p(\hat{\rvz}_{t-\tau+1:t}|\hat{\rvz}_{t-\tau})=\sum_{i=1}^n \log p(\hat{z}_{{t-\tau+1:t},i}|\hat{\rvz}_{t-\tau}).
	\end{align}
	By combining Equation (\ref{the7}) and Equation (\ref{the5}), we have:
	\begin{equation}
		\begin{split}
			\centering
			&p(h(\hat{\rvz}_{t-\tau+1:t})|\hat{\rvz}_{t-\tau})=p(\rvz_{t-\tau+1:t}|\rvz_{t-\tau}) \\ &\Longleftrightarrow p(\hat{\rvz}_{t-\tau+1:t}|\hat{\rvz}_{t-\tau})|\mathbf{J}_{h^{-1}}|=p(\rvz_{t-\tau+1:t}|\rvz_{t-\tau})\\&\Longleftrightarrow  \log p(\hat{z}_{{t-\tau+1:t}}|\hat{\rvz}_{t-\tau})=\sum_{k=1}^n \log p(z_{{t-\tau+1:t},k}|\rvz_{t-\tau}) - \log |\mathbf{J}_{h^{-1}}|,
		\end{split}
	\end{equation}
	where $\mathbf{J}_{h^{-1}}$ are the Jacobian matrix of $h^{-1}$.
	Sequentially, we take the first-order derivative with $\hat{\rvz}_{{t-\tau+1:t},i}$, where $i \in \{1,\cdots,n_s\}$ and have:
	\begin{align}
		\frac{\partial \log p(\hat{z}_{t-\tau+1:t}|\rvz^d_{t-\tau}) }{\partial \hat{z}^s_{t-\tau+1:t,i} } =  {\textstyle \sum_{k=1}^{n_s}}\frac{\partial \log{p(z^s_{t-\tau+1:t,k}|\rvz^d_{t-\tau})} }{\partial z^s_{t-\tau+1:t,k} } \cdot  \frac{\partial z^s_{t-\tau+1:t,k} }{\partial \hat{z}^s_{t-\tau+1:t,i}  } \\ \nonumber +
		{\textstyle \sum_{k=n_s+1}^{n}} \frac{\partial \log p(z^d_{t-\tau+1:t,k}|\rvz^s_{t-\tau})}{\partial z^d_{t-\tau+1:t,k} } \cdot \frac{\partial z^d_{t-\tau+1:t,k} }{\partial \hat{z}^s_{t-\tau+1:t,i}  } + \frac{\partial \log{|J_h|} }{\hat{z}^s_{t-\tau+1:t,i}} 
	\end{align}
	Then we further take the second-order derivative w.r.t $z^d_{t-\tau,l}$, where $l\in \{1,...,n_s\} $ and we have:
	\begin{align}
		\label{the8}
		\frac{\partial^2 \log p(\hat{z}_{t-\tau+1:t}|\rvz^d_{t-\tau}) }{\partial \hat{z}^s_{t-\tau+1:t,i} \partial z^d_{t-\tau,l}} =  {\textstyle \sum_{k=1}^{n_s}}\frac{\partial^2 \log{p(z^s_{t-\tau+1:t,k}|\rvz^d_{t-\tau})} }{\partial z^s_{t-\tau+1:t,k} \partial z^d_{t-\tau,l}} \cdot  \frac{\partial z^s_{t-\tau+1:t,k} }{\partial \hat{z}^s_{t-\tau+1:t,i}  } +
		\\
		\nonumber{\textstyle \sum_{k=n_s+1}^{n}} \frac{\partial^2 
			\log p(z^d_{t-\tau+1:t}|\rvz^d_{t-\tau})}{\partial z^d_{t-\tau+1:t,k} \partial z^d_{t-\tau,l}} \cdot \frac{\partial z^d_{t-\tau+1:t,k} }{\partial \hat{z}^s_{t-\tau+1:t,i}  } + \frac{\partial^2 \log{|J_h|} }{\partial \hat{z}^s_{t-\tau+1,l} \partial z^d_{t-\tau:t,i}} 
	\end{align}
	Since $\frac{\partial^2 \log p(\hat{z}_{t-\tau+1:t}|\rvz^d_{t-\tau}) }{\partial \hat{z}^s_{t-\tau+1:t,i} }$ does not change across different values of $z^d_{t-\tau,l}$, then $\frac{\partial^2 p(\hat{z}_{t-\tau+1:t}|\rvz^d_{t-\tau}) }{\partial \hat{z}^s_{t-\tau+1:t,i} \partial z^d_{t-\tau,l}}=0$. Moreover, since $\frac{\partial^2 \log{p(z^s_{t-\tau+1:t,k}|\rvz^d_{t-\tau})} }{\partial z^s_{t-\tau+1:t,k} \partial z^d_{t-\tau,l}} $ and $\frac{\partial^2 \log{|J_h|} }{\partial \hat{z}^s_{t-\tau+1,l} \partial z^d_{t-\tau:t,i}} =0$, Equation (\ref{the8}) can be further rewritten as:
	\begin{equation}
		\label{the9}
		{\textstyle \sum_{k=n_s+1}^{n}} \frac{\partial^2 \log p(z^d_{t-\tau+1:t}|\rvz^d_{t-\tau})}{\partial z^d_{t-\tau+1:t,k} \partial z^d_{t-\tau,l}} \cdot \frac{\partial z^d_{t-\tau+1:t,k} }{\partial \hat{z}^s_{t-\tau+1:t,i}  }=0
	\end{equation}
	By leveraging the linear independence assumption, the linear system denoted by Equation (\ref{the9}) has the only solution $\frac{\partial z^d_{t-\tau+1:t,k} }{\partial \hat{z}^s_{t-\tau+1:t,i}  }=0$. As $h$ is smooth, its Jacobian can written as:
	\begin{equation}  
		\begin{gathered}  
			\label{equ:Jh1}  
			\textbf{J}_{h_z}=\begin{bmatrix}  
				\begin{array}{c|c}  
					\textbf{A}:=\frac{\partial \rvz_t^{s}}{\partial \hat{\rvz}_t^{s}} & \textbf{B}:=\frac{\partial \rvz_t^s}{\partial \hat{\rvz}_t^{d}}=0 \\ \midrule  
					\textbf{C}:=\frac{\partial \rvz_t^{d}}{\partial \hat{\rvz}_t^s}=0 & \textbf{D}:=\frac{\partial \rvz_t^{d}}{\partial \hat{\rvz}_t^{d}}
				\end{array}  
			\end{bmatrix}  
		\end{gathered}  
	\end{equation} 
	Therefore, $\rvz_t^s$ and $\rvz_t^d$ are subspace identifiable.
\end{proof}

\subsection{Evident Lower Bound}
\label{app:elbo}
In this subsection, we show the evident lower bound. We first factorize the conditional distribution according to the Bayes theorem.
\begin{align}
	\ln p(\rvx_{1:H}) &= \ln \frac{p(\rvz^s_{1:H}, \rvz^d_{1:H}, \rvx_{1:H})}{p(\rvz^s_{1:H}, \rvz^d_{1:H}|\rvx_{1:H})} = \ln \frac{p(\rvx_{1:H}|\rvz^s_{1:H}, \rvz^d_{1:H})p(\rvz^s_{1:H}, \rvz^d_{1:H})}{p(\rvz^s_{1:H}, \rvz^d_{1:H}|\rvx_{1:H})}\\ \nonumber &= \mathbb{E}_{q(\rvz^s_{1:H},\rvz^d_{1:H}|\rvx_{1:H})} \ln \frac{p(\rvx_{1:H}|\rvz^s_{1:H}, \rvz^d_{1:H})p(\rvz^s_{1:H}, \rvz^d_{1:H})q(\rvz^s_{1:H}, \rvz^d_{1:H}|\rvx_{1:H})}{p(\rvz^s_{1:H}, \rvz^d_{1:H}|\rvx_{1:H})q(\rvz^s_{1:H}, \rvz^d_{1:H}|\rvz_{1:H})}\\
	\nonumber &\ge \mathbb{E}_{q(\rvz^s_{1:H},\rvz^d_{1:H}|\rvx_{1:H})}\ln p(\rvx_{1:H}|\rvz^s_{1:H},\rvz^d_{1:H}) -  D_{KL}(q(\rvz^s_{1:H},\rvz^d_{1:H}|\rvx_{1:H})||p(\rvz^s_{1:H},\rvz^d_{1:H}))  
\end{align}
specifically:
\begin{align}
	\ln \frac{p(\rvx_{1:H}|\rvz^s_{1:H}, \rvz^d_{1:H})p(\rvz^s_{1:H}, \rvz^d_{1:H})}{p(\rvz^s_{1:H}, \rvz^d_{1:H}|\rvx_{1:H})}&=\ln \frac{p(\rvx_{1:H}|\rvz^s_{1:H}, \rvz^d_{1:H})p(\rvz^s_{1:H}| \rvz^d_{1:H})p(\rvz^d_{1:H})}{p(\rvz^s_{1:H}, \rvz^d_{1:H}|\rvx_{1:H})}
\end{align}
\text{In the absence of $x_t$, $z^d_t$ and $z^s_t$ are independent. So the original expression is equal to:}
\begin{align}
	\nonumber
	&=\ln \frac{p(\rvx_{1:H}|\rvz^s_{1:H}, \rvz^d_{1:H})p(\rvz^s_{1:H})p(\rvz^d_{1:H})}{p(\rvz^s_{1:H}, \rvz^d_{1:H}|\rvx_{1:H})}=\ln \frac{p(\rvx_{1:H}|\rvz^s_{1:H}, \rvz^d_{1:H})p(\rvz^s_{1:H})p(\rvz^d_{1:H})q(\rvz^s_{1:H}|\rvx_{1:H})q(\rvz^d_{1:H}|\rvx_{1:H})}{p(\rvz^s_{1:H}, \rvz^d_{1:H}|\rvx_{1:H})q(\rvz^s_{1:H}|\rvx_{1:H})q(\rvz^d_{1:H}|\rvx_{1:H})}\\ \nonumber
	&\ge \mathbb{E}_{q(\rvz^s_{1:H}|\rvx_{1:H})} \mathbb{E}_{q(\rvz^d_{1:H}|\rvx_{1:H})}\ln p(\rvx_{1:H}|\rvz^s_{1:H},\rvz^d_{1:H}) -  D_{KL}(q(\rvz^s_{1:H}|\rvx_{1:H})||p(\rvz^s_{1:H}))\\
	\nonumber
	&-D_{KL}(q(\rvz^d_{1:H}|\rvx_{1:H})||p(\rvz^d_{1:H}))
\end{align}

\subsection{Prior Likelihood Derivation}
In this section, we derive the prior (As an example, for $z^s$) of $p(\hat{\rvz}_{1:t}^s)$ and $p(\hat{\rvz}_{1:t}^d)$ as follows:
We  consider the prior of $\ln p(\rvz_{1:t}^s)$. We start with an illustrative example of long/short-term latent causal processes with two time-delay latent variables, i.e. $\rvz^s_t=[z^s_{t,1}, z^s_{t,2}]$ with maximum time lag $L=1$, i.e., $z_{t,i}^s=f_i(\rvz_{t-1}^s, \epsilon_{t,i}^s)$ with mutually independent noises. Then we write this latent process as a transformation map $\mathbf{f}$ (note that we overload the notation $f$ for transition functions and for the transformation map):
\begin{equation}
	\small
	\begin{gathered}\nonumber
		\begin{bmatrix}
			\begin{array}{c}
				z_{t-1,1}^s \\ 
				z_{t-1,2}^s \\
				z_{t,1}^s   \\
				z_{t,2}^s
			\end{array}
		\end{bmatrix}=\mathbf{f}\left(
		\begin{bmatrix}
			\begin{array}{c}
				z_{t-1,1}^s \\ 
				z_{t-1,2}^s \\
				\epsilon_{t,1}^s   \\
				\epsilon_{t,2}^s
			\end{array}
		\end{bmatrix}\right).
	\end{gathered}
\end{equation}
By applying the change of variables formula to the map $\mathbf{f}$, we can evaluate the joint distribution of the latent variables $p(z_{t-1,1}^s,z_{t-1,2}^s,z_{t,1}^s, z_{t,2}^s)$ as 
\begin{equation}
	\small
	\label{equ:p1}
	p(z_{t-1,1}^s,z_{t-1,2}^s,z_{t,1}^s, z_{t,2}^s)=\frac{p(z_{t-1,1}^s, z_{t-1,2}^s, \epsilon_{t,1}^s, \epsilon_{t,2}^s)}{|\text{det }\mathbf{J}_{\mathbf{f}}|},
\end{equation}
where $\mathbf{J}_{\mathbf{f}}$ is the Jacobian matrix of the map $\mathbf{f}$, which is naturally a low-triangular matrix:
\begin{equation}
	\small
	\begin{gathered}\nonumber
		\mathbf{J}_{\mathbf{f}}=\begin{bmatrix}
			\begin{array}{cccc}
				1 & 0 & 0 & 0 \\
				0 & 1 & 0 & 0 \\
				\frac{\partial z_{t,1}^s}{\partial z_{t-1,1}^s} & \frac{\partial z_{t,1}^s}{\partial z_{t-1,2}^s} & 
				\frac{\partial z_{t,1}^s}{\partial \epsilon_{t,1}^s} & 0 \\
				\frac{\partial z_{t,2}^s}{\partial z_{t-1, 1}^s} &\frac{\partial z_{t,2}^s}{\partial z_{t-1,2}^s} & 0 & \frac{\partial z_{t,2}^s}{\partial \epsilon_{t,2}^s}
			\end{array}
		\end{bmatrix}.
	\end{gathered}
\end{equation}
Given that this Jacobian is triangular, we can efficiently compute its determinant as $\prod_i \frac{\partial z_{t,i}^s}{\epsilon_{t,i}^s}$. Furthermore, because the noise terms are mutually independent, and hence $\epsilon_{t,i}^s \perp \epsilon_{t,j}^s$ for $j\neq i$ and $\epsilon_{t}^s \perp \rvz_{t-1}^s$, so we can with the RHS of Equation (\ref{equ:p1}) as follows:
\begin{equation}
	\small
	\label{equ:p2}
	\begin{split}
		p(z_{t-1,1}^s, z_{t-1,2}^s, z_{t,1}^s, z_{t,2}^s)=p(z_{t-1,1}^s, z_{t-1,2}^s) \times \frac{p(\epsilon_{t,1}^s, \epsilon_{t,2}^s)}{|\mathbf{J}_{\mathbf{f}}|}=p(z_{t-1,1}^s, z_{t-1,2}^s) \times \frac{\prod_i p(\epsilon_{t,i}^s)}{|\mathbf{J}_{\mathbf{f}}|}.
	\end{split}
\end{equation}
Finally, we generalize this example and derive the prior likelihood below. Let $\{r_i^s\}_{i=1,2,3,\cdots}$ be a set of learned inverse transition functions that take the estimated latent causal variables, and output the noise terms, i.e., $\hat{\epsilon}_{t,i}^s=r_i^s(\hat{z}_{t,i}^s, \{ \hat{\rvz}_{t-\tau}^s\})$. Then we design a transformation $\mathbf{A}\rightarrow \mathbf{B}$ with low-triangular Jacobian as follows:
\begin{equation}
	\small
	\begin{gathered}
		\underbrace{[\hat{\rvz}_{t-L}^s,\cdots,{\hat{\rvz}}_{t-1}^s,{\hat{\rvz}}_{t}^s]^{\top}}_{\mathbf{A}} \text{  mapped to  } \underbrace{[{\hat{\rvz}}_{t-L}^s,\cdots,{\hat{\rvz}}_{t-1}^s,{\hat{\epsilon}}_{t,i}^s]^{\top}}_{\mathbf{B}}, \text{ with } \mathbf{J}_{\mathbf{A}\rightarrow\mathbf{B}}=
		\begin{bmatrix}
			\begin{array}{cc}
				\mathbb{I}_{n_s\times L} & 0\\
				* & \text{diag}\left(\frac{\partial r^s_{i,j}}{\partial {\hat{z}}^s_{t,j}}\right)
			\end{array}
		\end{bmatrix}.
	\end{gathered}
\end{equation}
Similar to Equation (\ref{equ:p2}), we can obtain the joint distribution of the estimated dynamics subspace as:
\begin{equation}
	\log p(\mathbf{A})=\underbrace{\log p({\hat{\rvz}}^s_{t-L},\cdots, {\hat{\rvz}}^s_{t-1}) + \sum^{n_s}_{i=1}\log p({\hat{\epsilon}}^s_{t,i})}_{\text{Because of mutually independent noise assumption}}+\log (|\text{det}(\mathbf{J}_{\mathbf{A}\rightarrow\mathbf{B}})|)
\end{equation}
Finally, we have:
\begin{equation}
	\small
	\log p({\hat{\rvz}}_t^s|\{{\hat{\rvz}}_{t-\tau}^s\}_{\tau=1}^L)=\sum_{i=1}^{n_s}\log p({\hat{\epsilon}_{t,i}^s}) + \sum_{i=1}^{n_s}\log |\frac{\partial r^s_i}{\partial {\hat{z}}^s_{t,i}}|
\end{equation} 
Since the prior of $p(\hat{\rvz}_{t+1:T}^s|\hat{\rvz}_{1:t}^s)=\prod_{i=t+1}^{T} p(\hat{\rvz}_{i}^s|\hat{\rvz}_{i-1}^s)$ with the assumption of first-order Markov assumption, we can estimate $p(\hat{\rvz}_{t+1:T}^s|\hat{\rvz}_{1:t}^s)$ in a similar way.

\section{C. Implementation Details}

\subsection{Model Details}
We choose FSNet \cite{pham2022learning} as the encoder backbone of long latent variables and MLP as the encoder backbone of short latent variables. Specifically, 
given the FSNet and MLP extract the hidden feature, $z^d_t \text{ and } z^s_t$,  we apply a variational inference block and then a 
MLP-based decoder. Meanwhile, inspired by OneNet \cite{wen2024onenet}, we introduce a hyperparameter “mode” to choose between feature-based and time-based to extraction extract the hidden feature. Architecture details of the proposed method are shown in Table\ref{config}

\begin{table}[]
	
	\caption{Architecture details. H: length of time series, L: length of observed time series, LeakyReLU: Leaky Rectified
		Linear Unit, $|x_t|:$ the dimension of $x_t$}
	\label{config}
	\renewcommand{\arraystretch}{1.2}
	\setlength{\tabcolsep}{5pt}
	\begin{tabular}{cl|cl|cl}
		\hline
		\multicolumn{2}{c|}{Configuration}                          & \multicolumn{2}{c|}{Description}                        & \multicolumn{2}{c}{Output}                                                                                 \\ \hline
		\multicolumn{2}{c|}{$1.\phi ^ s$}                           & \multicolumn{2}{c|}{Long-term Variable Encoder}         & \multicolumn{2}{c}{}                                                                                       \\ \hline
		\multicolumn{2}{c|}{Input $\rvx_{1:L}$}                     & \multicolumn{2}{c|}{Observed time series}               & \multicolumn{2}{c}{$Batch Size \times L \times |\rvx_t|$}                                                  \\
		\multicolumn{2}{c|}{Dense}                                  & \multicolumn{2}{c|}{Conv1d}                             & \multicolumn{2}{c}{$Batch Size \times 640 \times |\rvx_t|$}                                                \\
		\multicolumn{2}{c|}{Dense}                                  & \multicolumn{2}{c|}{L neurons; LeakyReLU}               & \multicolumn{2}{c}{$Batch Size \times L \times |\rvx_t|$}                                                  \\ \hline
		\multicolumn{2}{c|}{$2.\phi ^ d$}                           & \multicolumn{2}{c|}{Short-term Variable Encoder}        & \multicolumn{2}{c}{}                                                                                       \\ \hline
		\multicolumn{2}{c|}{Input $\rvx_{1:L}$}                     & \multicolumn{2}{c|}{Observed time series}               & \multicolumn{2}{c}{$Batch Size \times L \times |\rvx_t|$}                                                  \\
		\multicolumn{2}{c|}{Dense}                                  & \multicolumn{2}{c|}{512 neurons; LeakyReLU}             & \multicolumn{2}{c}{$Batch Size \times 512 \times |\rvx_t|$}                                                \\
		\multicolumn{2}{c|}{Dense}                                  & \multicolumn{2}{c|}{L neurons; LeakyReLU}               & \multicolumn{2}{c}{$Batch Size \times L \times |\rvx_t|$}                                                  \\ \hline
		\multicolumn{2}{c|}{$3.T^s$}                                & \multicolumn{2}{c|}{Long-term Variable Predict Module}  & \multicolumn{2}{c}{}                                                                                       \\ \hline
		\multicolumn{2}{c|}{Input $\rvz^s_{1:L}$}                   & \multicolumn{2}{c|}{Long-term Latent Variables}         & \multicolumn{2}{c}{$Batch Size \times L \times |\rvx_t|$}                                                  \\
		\multicolumn{2}{c|}{Dense}                                  & \multicolumn{2}{c|}{512 neurons; LeakyReLU}             & \multicolumn{2}{c}{$Batch Size \times 512  \times |\rvx_t|$}                                               \\
		\multicolumn{2}{c|}{Dense}                                  & \multicolumn{2}{c|}{(H-L) neurons; LeakyReLU}           & \multicolumn{2}{c}{$Batch Size \times (H-L)  \times |\rvx_t|$}                                             \\ \hline
		\multicolumn{2}{c|}{$4.T^d$}                                & \multicolumn{2}{c|}{Short-term Variable Predict Module} & \multicolumn{2}{c}{}                                                                                       \\ \hline
		\multicolumn{2}{c|}{Input $\rvz^d_{1:L}$}                   & \multicolumn{2}{c|}{Long-term Latent Variables}         & \multicolumn{2}{c}{$Batch Size \times L \times |\rvx_t|$}                                                  \\
		\multicolumn{2}{c|}{Dense}                                  & \multicolumn{2}{c|}{(H-L) neurons; LeakyReLU}           & \multicolumn{2}{c}{$Batch Size \times (H-L)  \times |\rvx_t|$}                                             \\ \hline
		\multicolumn{2}{c|}{$5.F_x$}                                & \multicolumn{2}{c|}{Historical Decoder}                 & \multicolumn{2}{c}{}                                                                                       \\ \hline
		\multicolumn{2}{c|}{Input $\rvz^s_{1:L}, \rvz^d_{1:L}$}     & \multicolumn{2}{c|}{Long/Short-term Latent Variable}    & \multicolumn{2}{c}{$Batch Size \times L  \times |\rvx_t|$ ,$Batch Size \times L  \times |\rvx_t|$}         \\
		\multicolumn{2}{c|}{Concat}                                 & \multicolumn{2}{c|}{concatenation}                      & \multicolumn{2}{c}{$Batch Size \times L  \times 2|\rvx_t|$}                                                \\
		\multicolumn{2}{c|}{Dense}                                  & \multicolumn{2}{c|}{$|\rvx_t|$ neurons; LeakyReLU}      & \multicolumn{2}{c}{$Batch Size \times L \times |\rvx_t|$}                                                  \\ \hline
		\multicolumn{2}{c|}{$6.F_y$}                                & \multicolumn{2}{c|}{Future Predictor}                   & \multicolumn{2}{c}{}                                                                                       \\ \hline
		\multicolumn{2}{c|}{Input $\rvz^s_{L+1:H}, \rvz^d_{L+1:H}$} & \multicolumn{2}{c|}{Long/Short-term Latent Variable}    & \multicolumn{2}{c}{$Batch Size \times (H-L)  \times |\rvx_t|$ ,$Batch Size \times (H-L)  \times |\rvx_t|$} \\
		\multicolumn{2}{c|}{Concat}                                 & \multicolumn{2}{c|}{concatenation}                      & \multicolumn{2}{c}{$Batch Size \times (H-L) \times 2|\rvx_t|$}                                             \\
		\multicolumn{2}{c|}{Dense}                                  & \multicolumn{2}{c|}{512 neurons; LeakyReLU}             & \multicolumn{2}{c}{$Batch Size \times (H-L) \times 512$}                                                   \\
		\multicolumn{2}{c|}{Dense}                                  & \multicolumn{2}{c|}{$|\rvx_t|$  neurons; LeakyReLU}     & \multicolumn{2}{c}{$Batch Size \times (H-L) \times |\rvx_t|$}                                              \\ \hline
		\multicolumn{2}{c|}{$7.r$}                                  & \multicolumn{2}{c|}{Modular Prior Networks}             & \multicolumn{2}{c}{}                                                                                       \\ \hline
		\multicolumn{2}{c|}{Input $\rvz^s_{1:L} or \rvz^d_{1:L}$}   & \multicolumn{2}{c|}{Latent Variables}                   & \multicolumn{2}{c}{$Batch Size \times (n_*+1)$}                                                            \\
		\multicolumn{2}{c|}{Dense}                                  & \multicolumn{2}{c|}{128 neurons,LeakyReLU}              & \multicolumn{2}{c}{$(n_*+1) \times 128$}                                                                   \\
		\multicolumn{2}{c|}{Dense}                                  & \multicolumn{2}{c|}{128 neurons,LeakyReLU}              & \multicolumn{2}{c}{$128 \times 128$}                                                                       \\
		\multicolumn{2}{c|}{Dense}                                  & \multicolumn{2}{c|}{128 neurons,LeakyReLU}              & \multicolumn{2}{c}{$128 \times 128$}                                                                       \\
		\multicolumn{2}{c|}{JacobianCompute}                        & \multicolumn{2}{c|}{Compute log (det (J))}            & \multicolumn{2}{c}{$Batch Size$}                                                                             \\ \hline
	\end{tabular}%
	
\end{table}
\subsection{Experiment Details.}
We use ADAM optimizer in all experiments and report the mean squared error (MSE) and the 
mean absolute error (MAE) as evaluation metrics. All experiments are implemented by Pytorch on a 
single NVIDIA GTX 3090 24GB GPU.

\section{D. More Experiments Results}

\begin{figure}
	\centering
	\includegraphics[width=1\linewidth]{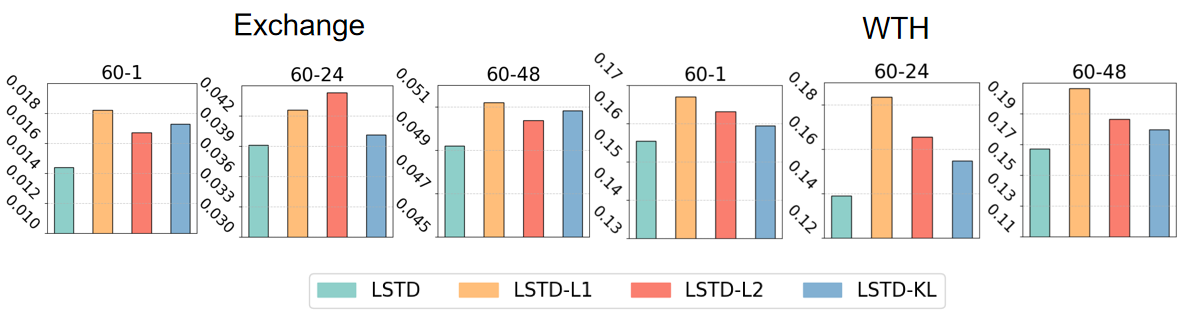}
	\caption{All ablation experiments}
	\label{fig:all_abl}
\end{figure}

\subsection{Ablation Study}
In the main text, we conducted three ablation experiments. All ablation experiments are shown in the figure \ref{fig:all_abl}.It can be shown in Table \ref{tab:few samples} that our model can adapt to the nonstationary data with few samples. We can find that even if the input horizon is small, e.g. 20, our method can successfully adapt to the nonstationary data with few samples. 

\begin{table}[]
	\caption{Experiments results on the Exchange data with different input horizons}
	\centering
	\label{tab:few samples}
	\begin{tabular}{c|cc|cc|cc|cc|cc}
		\hline
		Length   & \multicolumn{2}{c|}{20-24}      & \multicolumn{2}{c|}{30-24}      & \multicolumn{2}{c|}{40-24}      & \multicolumn{2}{c|}{50-24}      & \multicolumn{2}{c}{60-24}       \\ \hline
		Method   & MSE            & MAE            & MSE            & MAE            & MSE            & MAE            & MSE            & MAE            & MSE            & MAE            \\ \hline
		LSTD     & \textbf{0.043} & \textbf{0.138} & \textbf{0.046} & \textbf{0.144} & \textbf{0.044} & \textbf{0.141} & \textbf{0.039} & \textbf{0.132} & \textbf{0.039} & \textbf{0.132} \\
		OneNet   & 0.048          & 0.148          & 0.049          & 0.152          & 0.048          & 0.147          & 0.049          & 0.15           & 0.047          & 0.148          \\
		FSNet    & 0.133          & 0.223          & 0.125          & 0.210           & 0.141          & 0.216          & 0.125          & 0.208          & 0.113          & 0.206          \\
		OneNet-T & 0.063          & 0.168          & 0.059          & 0.164          & 0.059          & 0.164          & 0.604          & 0.166          & 0.060           & 0.166          \\
		DER++    & 0.138          & 0.231          & 0.127          & 0.237          & 0.149          & 0.234          & 0.121          & 0.229          & 0.111          & 0.227          \\
		ER       & 0.183          & 0.218          & 0.178          & 0.22           & 0.188          & 0.225          & 0.174          & 0.212          & 0.162          & 0.210           \\
		MIR      & 0.122          & 0.204          & 0.115          & 0.210           & 0.119          & 0.207          & 0.113          & 0.207          & 0.104          & 0.204          \\
		TFCL     & 0.106          & 0.239          & 0.110           & 0.246          & 0.106          & 0.233          & 0.103          & 0.241          & 0.098          & 0.227          \\
		Online-T & 0.134          & 0.217          & 0.128          & 0.231          & 0.137          & 0.224          & 0.122          & 0.221          & 0.116          & 0.213          \\
		Informer & 0.123          & 0.237          & 0.134          & 0.231          & 0.117          & 0.229          & 0.121          & 0.224          & 0.107          & 0.196          \\ \hline
	\end{tabular}
\end{table}

\subsection{Model Efficiency}
\begin{figure}  
	\centering  
	\includegraphics[width=0.5\linewidth]{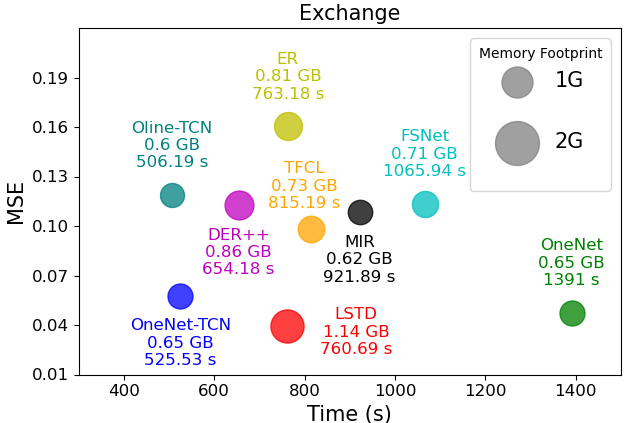}  
	\caption{Model efficiency comparison of different methods. Training time and memory footprint are recorded with the same batch size and official configuration.}  
	\label{fig:efficiency}  
\end{figure}  
We evaluate the performance of our model and the baseline model on the exchange rate dataset from three aspects: forecasting
performance, training speed, and memory footprint as shown in Figure \ref{fig:efficiency}. Compared with other models for online time-series forecasting, we can find that the proposed \textbf{LSTD} has the best model performance and relatively good model efficiency. However, in terms of memory performance, \textbf{LSTD} may be lower than other baselines because we need to incorporate priors during the training process. 

\subsection{Experiments of Real-world Datasets}
\begin{figure}
	\centering
	\includegraphics[width=1\linewidth]{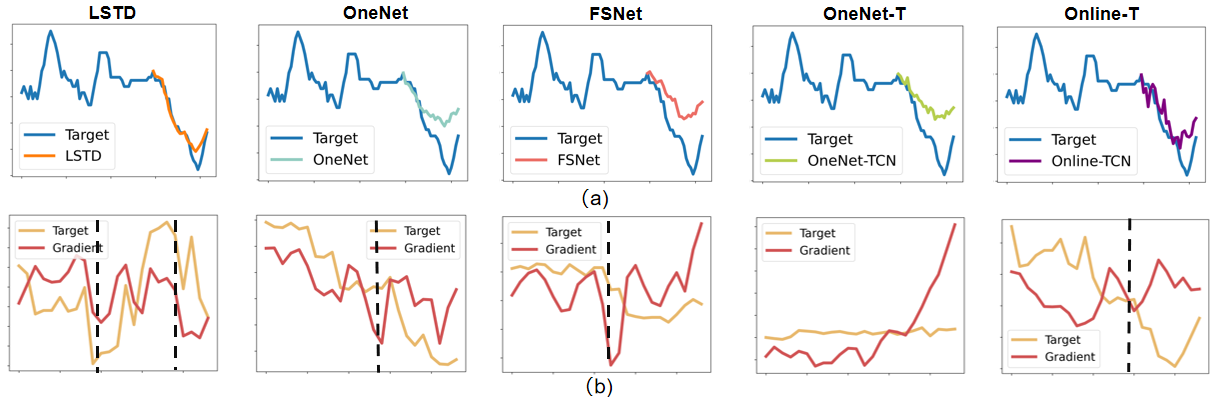}
	\caption{The figure  represents visualization of the proposed LSTD and other baselines. The blue lines denote the ground-truth time series data and the lines with other colors denote the predicted results of different methods.}
	\label{fig:predict}
\end{figure}

In the main text, to ensure the reproducibility of other baselines, we used the tabular data from the OneNet\cite{wen2024onenet}, For the sake of fairness, we also ran the experiment three times on the same machine with seeds 2023, 2024, 2025. And we recorded the mean and variance of all methods. The results can be seen in Table \ref{MSE_our} to Table \ref{MAE_std}. The complete visualization in the Weather dataset is shown in Figure \ref{fig:predict}.
\begin{table}[]
	\caption{Mean Square Error (MSE) results on the different datasets with three different seeds. TCN is abbreviated as T}
	\label{MSE_our}
	\resizebox{\textwidth}{!}{%
		\begin{tabular}{c|c|cccccccccc}
			\hline
			Models                     & Len & LSTD                                  & OneNet         & FSNet & OneNet-T & DER++ & ER    & MIR   & TFCL   & Online-T & Informer \\ \hline
			& 1   & \textbf{0.375}                        & 0.377          & 0.491 & 0.394    & 0.511 & 0.581 & 0.524 & 0.531  & 0.617    & 8.126    \\
			& 24  & \textbf{0.543}                        & 0.548          & 0.768 & 0.943    & 0.836 & 0.802 & 0.816 & 0.851  & 0.832    & 5.173    \\
			\multirow{-3}{*}{ETTh2}    & 48  & \textbf{0.616}                        & 0.622          & 0.948 & 0.926    & 1.157 & 1.103 & 1.098 & 1.211  & 1.188    & 6.272    \\ \hline
			& 1   & \textbf{0.082}                        & 0.086          & 0.093 & 0.091    & 0.082 & 0.089 & 0.082 & 0.085  & 0.208    & 0.443    \\
			& 24  & \textbf{0.102}                        & 0.105          & 0.195 & 0.213    & 0.198 & 0.205 & 0.189 & 0.216  & 0.263    & 0.465    \\
			\multirow{-3}{*}{ETTm1}    & 48  & 0.115                                 & \textbf{0.110} & 0.185 & 0.216    & 0.203 & 0.231 & 0.223 & 0.240  & 0.271    & 0.391    \\ \hline
			& 1   & \textbf{0.155}                        & 0.157          & 0.167 & 0.157    & 0.172 & 0.182 & 0.182 & 0.176  & 0.213    & 0.422    \\
			& 24  & \textbf{0.139}                        & 0.173          & 0.177 & 0.276    & 0.279 & 0.290 & 0.286 & 0.311  & 0.312    & 0.376    \\
			\multirow{-3}{*}{WTH}      & 48  & \textbf{0.167}                        & 0.196          & 0.193 & 0.289    & 0.233 & 0.301 & 0.289 & 0.324  & 0.298    & 0.384    \\ \hline
			& 1   & \textbf{2.228}                        & 2.675          & 3.586 & 2.413    & 2.692 & 2.618 & 2.568 & 2.806  & 3.312    & 3.813    \\
			& 24  & \textbf{1.557}                        & 2.090          & 6.055 & 4.551    & 8.961 & 9.235 & 9.157 & 11.891 & 11.594   & 9.185    \\
			\multirow{-3}{*}{ECL}      & 48  & \textbf{1.720}                        & 2.438          & 7.881 & 4.488    & 8.994 & 9.597 & 9.391 & 12.109 & 11.912   & 10.183   \\ \hline
			& 1   & \textbf{0.234}                        & 0.241          & 0.312 & 0.236    & 0.271 & 0.284 & 0.298 & 0.306  & 0.334    & 0.234    \\
			& 24  & \textbf{0.417}                        & 0.438          & 0.426 & 0.425    & 0.476 & 0.461 & 0.451 & 0.441  & 0.481    & 0.451    \\
			\multirow{-3}{*}{Traffic}  & 48  & \textbf{0.431}                        & 0.473          & 0.445 & 0.451    & 0.486 & 0.510 & 0.502 & 0.438  & 0.503    & 0.496    \\ \hline
			& 1   & {\color[HTML]{212121} \textbf{0.014}} & 0.017          & 0.094 & 0.031    & 0.106 & 0.097 & 0.095 & 0.106  & 0.113    & 0.102    \\
			& 24  & \textbf{0.039}                        & 0.047          & 0.113 & 0.060    & 0.111 & 0.162 & 0.104 & 0.098  & 0.116    & 0.107    \\
			\multirow{-3}{*}{Exchange} & 48  & \textbf{0.049}                        & 0.062          & 0.156 & 0.065    & 0.183 & 0.181 & 0.101 & 0.101  & 0.168    & 0.116    \\ \hline
		\end{tabular}%
	}
\end{table}
\begin{table}[]
	\caption{Mean Absolute Error (MAE) results on the different datasets with three different seeds. TCN is abbreviated as T}
	\label{MAE_our}
	\resizebox{\textwidth}{!}{%
		\begin{tabular}{c|c|cccccccccc}
			\hline
			Models                     & Len & LSTD                                  & OneNet         & FSNet & OneNet-T & DER++ & ER    & MIR   & TFCL  & Online-T & Informer \\ \hline
			& 1   & \textbf{0.347}                        & 0.354          & 0.471 & 0.373    & 0.373 & 0.381 & 0.418 & 0.466 & 0.443    & 0.852    \\
			& 24  & \textbf{0.411}                        & 0.415          & 0.557 & 0.532    & 0.536 & 0.515 & 0.543 & 0.539 & 0.545    & 0.652    \\
			\multirow{-3}{*}{ETTh2}    & 48  & \textbf{0.423}                        & 0.448          & 0.541 & 0.535    & 0.579 & 0.586 & 0.572 & 0.591 & 0.598    & 0.768    \\ \hline
			& 1   & \textbf{0.189}                        & 0.192          & 0.244 & 0.207    & 0.196 & 0.208 & 0.201 & 0.192 & 0.218    & 0.510    \\
			& 24  & \textbf{0.217}                        & 0.234          & 0.326 & 0.343    & 0.331 & 0.343 & 0.327 & 0.346 & 0.376    & 0.518    \\
			\multirow{-3}{*}{ETTm1}    & 48  & 0.259                                 & \textbf{0.242} & 0.319 & 0.348    & 0.356 & 0.367 & 0.347 & 0.357 & 0.415    & 0.462    \\ \hline
			& 1   & \textbf{0.200}                        & 0.202          & 0.224 & 0.206    & 0.176 & 0.182 & 0.182 & 0.169 & 0.210    & 0.438    \\
			& 24  & \textbf{0.224}                        & 0.255          & 0.261 & 0.337    & 0.304 & 0.284 & 0.317 & 0.314 & 0.317    & 0.380    \\
			\multirow{-3}{*}{WTH}      & 48  & \textbf{0.250}                        & 0.277          & 0.276 & 0.354    & 0.300 & 0.296 & 0.289 & 0.325 & 0.334    & 0.361    \\ \hline
			& 1   & \textbf{0.232}                        & 0.268          & 0.545 & 0.280    & 0.428 & 0.518 & 0.519 & 0.273 & 0.641    & 0.549    \\
			& 24  & \textbf{0.288}                        & 0.341          & 1.116 & 0.405    & 1.081 & 1.184 & 1.035 & 1.194 & 1.291    & 1.098    \\
			\multirow{-3}{*}{ECL}      & 48  & \textbf{0.348}                        & 0.367          & 1.194 & 0.423    & 1.054 & 1.031 & 1.184 & 1.304 & 1.219    & 1.164    \\ \hline
			& 1   & \textbf{0.229}                        & 0.240          & 0.278 & 0.236    & 0.251 & 0.256 & 0.284 & 0.297 & 0.284    & 0.258    \\
			& 24  & \textbf{0.332}                        & 0.346          & 0.365 & 0.346    & 0.409 & 0.417 & 0.443 & 0.493 & 0.385    & 0.365    \\
			\multirow{-3}{*}{Traffic}  & 48  & \textbf{0.344}                        & 0.371          & 0.378 & 0.355    & 0.386 & 0.294 & 0.397 & 0.531 & 0.380    & 0.394    \\ \hline
			& 1   & {\color[HTML]{212121} \textbf{0.070}} & 0.085          & 0.174 & 0.117    & 0.173 & 0.124 & 0.118 & 0.153 & 0.169    & 0.115    \\
			& 24  & \textbf{0.132}                        & 0.148          & 0.206 & 0.166    & 0.227 & 0.210 & 0.204 & 0.227 & 0.213    & 0.196    \\
			\multirow{-3}{*}{Exchange} & 48  & \textbf{0.150}                        & 0.170          & 0.254 & 0.173    & 0.243 & 0.241 & 0.209 & 0.183 & 0.258    & 0.217    \\ \hline
		\end{tabular}%
	}
\end{table}

\begin{table}[]
	\caption{Standard deviation of MSE results on the different datasets}
	\label{MSE_std}
	\renewcommand{\arraystretch}{1.1}
	\resizebox{\textwidth}{!}{%
		\begin{tabular}{c|c|cccccccccc}
			\hline
			Models                     & Len & LSTD                            & OneNet   & FSNet    & OneNet-T & DER++    & ER       & MIR      & TFCL     & Online-T & Informer \\ \hline
			& 1   & 6.05e-04                        & 7.29e-05 & 3.43e-01 & 5.74e-05 & 2.65e-03 & 1.72e-02 & 1.51e-02 & 3.33e-02 & 2.23e-04 & 2.59e-03 \\
			& 24  & 6.79e-04                        & 1.87e-04 & 7.85e-01 & 2.44e-02 & 2.21e-02 & 3.17e-03 & 2.62e-02 & 4.89e-03 & 9.52e-05 & 1.52e-03 \\
			\multirow{-3}{*}{ETTh2}    & 48  & 8.70e-04                        & 6.92e-04 & 8.84e-03 & 6.87e-03 & 1.45e-02 & 2.74e-03 & 6.84e-03 & 1.74e-02 & 2.94e-04 & 1.29e-03 \\ \hline
			& 1   & 9.35e-07                        & 6.26e-06 & 1.69e-03 & 7.26e-05 & 1.75e-03 & 8.90e-03 & 1.91e-02 & 1.58e-02 & 8.73e-05 & 1.71e-04 \\
			& 24  & 1.05e-07                        & 3.10e-06 & 6.37e-04 & 6.10e-06 & 1.75e-02 & 2.69e-02 & 1.23e-02 & 2.70e-02 & 6.35e-05 & 2.52e-04 \\
			\multirow{-3}{*}{ETTm1}    & 48  & 2.46e-04                        & 1.45e-06 & 4.66e-04 & 4.67e-05 & 1.27e-02 & 2.44e-02 & 1.37e-03 & 5.83e-03 & 2.01e-04 & 1.43e-03 \\ \hline
			& 1   & 7.15e-07                        & 3.81e-07 & 8.82e-07 & 2.38e-07 & 1.44e-02 & 2.19e-02 & 3.05e-02 & 1.54e-02 & 2.61e-04 & 1.86e-04 \\
			& 24  & 5.45e-06                        & 6.76e-06 & 3.50e-06 & 5.44e-06 & 1.01e-02 & 1.18e-02 & 1.47e-02 & 3.18e-02 & 2.78e-04 & 3.02e-04 \\
			\multirow{-3}{*}{WTH}      & 48  & 3.07e-05                        & 1.69e-05 & 1.11e-05 & 1.01e-04 & 1.14e-03 & 2.15e-02 & 1.79e-02 & 3.20e-02 & 1.17e-04 & 8.04e-05 \\ \hline
			& 1   & 3.87e-04                        & 2.02e-03 & 3.17e-02 & 1.18e-03 & 2.90e-03 & 3.06e-04 & 1.42e-04 & 6.65e-03 & 2.32e-04 & 1.60e-03 \\
			& 24  & 6.57e-04                        & 4.19e-04 & 2.39e-01 & 1.06e-03 & 1.43e-04 & 1.67e-04 & 2.56e-05 & 1.07e-04 & 3.17e-04 & 3.25e-03 \\
			\multirow{-3}{*}{ECL}      & 48  & 4.26e-02                        & 1.00e-02 & 4.15e-01 & 2.30e-04 & 2.27e-02 & 3.27e-04 & 1.58e-04 & 6.49e-05 & 8.39e-05 & 1.54e-04 \\ \hline
			& 1   & 7.33e-06                        & 1.06e-06 & 9.96e-06 & 2.30e-06 & 1.14e-04 & 6.80e-05 & 2.21e-04 & 2.67e-05 & 2.14e-03 & 2.42e-04 \\
			& 24  & 4.89e-05                        & 4.59e-05 & 2.70e-05 & 2.13e-05 & 3.87e-03 & 1.35e-04 & 2.18e-04 & 1.06e-04 & 1.80e-04 & 1.19e-04 \\
			\multirow{-3}{*}{Traffic}  & 48  & 5.93e-06                        & 8.38e-04 & 6.02e-05 & 2.22e-06 & 9.09e-05 & 1.16e-04 & 2.52e-04 & 3.05e-04 & 3.23e-04 & 6.74e-05 \\ \hline
			& 1   & {\color[HTML]{212121} 2.85e-07} & 1.22e-06 & 8.55e-04 & 4.23e-06 & 1.68e-05 & 3.50e-05 & 1.69e-04 & 2.49e-04 & 7.35e-05 & 2.12e-04 \\
			& 24  & 1.66e-08                        & 4.10e-05 & 2.13e-04 & 9.82e-07 & 5.92e-05 & 2.40e-04 & 4.92e-05 & 1.92e-04 & 1.88e-04 & 5.37e-05 \\
			\multirow{-3}{*}{Exchange} & 48  & 2.95e-06                        & 2.91e-04 & 1.35e-04 & 4.79e-06 & 7.38e-05 & 4.58e-05 & 1.47e-05 & 1.59e-04 & 4.48e-05 & 1.81e-04 \\ \hline
		\end{tabular}%
	}
\end{table}

\begin{table}[]
	\caption{Standard deviation of MAE results on the different datasets}
	\label{MAE_std}
	\renewcommand{\arraystretch}{1.1}
	\resizebox{\textwidth}{!}{%
		\begin{tabular}{c|c|cccccccccc}
			\hline
			Models                     & Len & LSTD                            & OneNet   & FSNet    & OneNet-T & DER++    & ER       & MIR      & TFCL     & Online-T & Informer \\ \hline
			& 1   & 5.38e-05                        & 2.86e-05 & 3.38e-04 & 4.41e-05 & 1.96e-03 & 7.04e-05 & 1.70e-04 & 3.34e-04 & 3.08e-04 & 1.26e-04 \\
			& 24  & 9.95e-06                        & 3.71e-05 & 1.23e-03 & 3.02e-04 & 1.92e-03 & 2.35e-04 & 1.66e-03 & 2.21e-03 & 4.39e-05 & 3.15e-05 \\
			\multirow{-3}{*}{ETTh2}    & 48  & 3.66e-05                        & 9.72e-05 & 5.46e-05 & 1.56e-04 & 1.92e-04 & 3.07e-05 & 9.69e-05 & 1.35e-05 & 1.33e-04 & 6.90e-05 \\ \hline
			& 1   & 4.62e-06                        & 2.06e-05 & 1.61e-03 & 1.51e-04 & 3.02e-03 & 1.63e-04 & 2.05e-04 & 2.67e-03 & 3.06e-04 & 2.17e-04 \\
			& 24  & 6.97e-08                        & 4.91e-06 & 5.10e-04 & 4.02e-06 & 2.23e-03 & 6.26e-05 & 2.76e-03 & 3.26e-04 & 2.27e-05 & 6.31e-05 \\
			\multirow{-3}{*}{ETTm1}    & 48  & 1.38e-04                        & 1.44e-06 & 4.31e-04 & 3.80e-05 & 1.23e-03 & 6.94e-05 & 7.37e-05 & 3.19e-03 & 1.59e-04 & 2.96e-04 \\ \hline
			& 1   & 1.11e-06                        & 2.58e-07 & 1.90e-06 & 2.00e-07 & 2.30e-03 & 2.60e-05 & 8.26e-05 & 2.65e-04 & 6.14e-05 & 1.28e-03 \\
			& 24  & 1.07e-06                        & 7.48e-06 & 4.08e-06 & 5.36e-06 & 3.05e-03 & 1.06e-04 & 6.12e-05 & 8.00e-05 & 2.46e-04 & 2.12e-06 \\
			\multirow{-3}{*}{WTH}      & 48  & 1.94e-05                        & 1.50e-05 & 8.51e-06 & 2.48e-05 & 3.03e-03 & 2.99e-04 & 1.75e-04 & 3.38e-04 & 4.01e-05 & 2.84e-04 \\ \hline
			& 1   & 3.05e-05                        & 1.58e-06 & 2.62e-04 & 1.99e-05 & 9.00e-04 & 1.12e-04 & 7.17e-05 & 3.30e-03 & 1.63e-06 & 2.16e-04 \\
			& 24  & 1.91e-05                        & 6.26e-06 & 1.79e-04 & 1.20e-08 & 1.36e-03 & 2.35e-05 & 3.38e-03 & 3.09e-04 & 2.31e-04 & 3.89e-05 \\
			\multirow{-3}{*}{ECL}      & 48  & 4.71e-03                        & 2.19e-05 & 1.95e-04 & 1.60e-06 & 1.20e-03 & 5.26e-05 & 1.56e-04 & 3.16e-05 & 3.21e-04 & 1.78e-03 \\ \hline
			& 1   & 7.73e-06                        & 3.32e-07 & 7.15e-06 & 2.02e-06 & 7.03e-04 & 1.49e-04 & 5.18e-05 & 1.56e-04 & 1.11e-04 & 2.21e-04 \\
			& 24  & 1.79e-05                        & 7.31e-06 & 1.71e-05 & 7.16e-06 & 3.78e-04 & 1.88e-04 & 1.13e-04 & 2.25e-04 & 3.30e-04 & 7.38e-05 \\
			\multirow{-3}{*}{Traffic}  & 48  & 4.36e-06                        & 2.08e-04 & 3.21e-05 & 1.72e-06 & 2.92e-04 & 2.11e-04 & 7.86e-06 & 2.71e-04 & 2.79e-06 & 1.60e-04 \\ \hline
			& 1   & {\color[HTML]{212121} 3.24e-06} & 9.36e-06 & 6.65e-04 & 1.13e-05 & 1.34e-05 & 2.93e-05 & 8.01e-05 & 1.01e-06 & 2.97e-04 & 8.06e-05 \\
			& 24  & 1.66e-08                        & 9.11e-05 & 8.26e-05 & 1.93e-06 & 1.26e-05 & 1.12e-04 & 1.94e-04 & 2.59e-05 & 1.55e-06 & 2.13e-04 \\
			\multirow{-3}{*}{Exchange} & 48  & 1.06e-05                        & 5.79e-04 & 1.08e-04 & 9.63e-06 & 8.39e-06 & 1.86e-04 & 1.86e-06 & 3.01e-04 & 3.05e-08 & 3.42e-04 \\ \hline
		\end{tabular}%
	}
\end{table}